\documentclass{article}
\usepackage{newtxtext}
\usepackage{microtype}

\usepackage[preprint]{main}
\usepackage[utf8]{inputenc} 
\usepackage[T1]{fontenc}    
\usepackage{hyperref}       
\usepackage{url}            
\usepackage{booktabs}       
\usepackage{amsfonts}       
\usepackage{nicefrac}       
\usepackage{microtype}      
\usepackage{xcolor}         
\usepackage{listings}

\usepackage{graphicx}
\usepackage{multirow} 
\usepackage{enumitem}
\usepackage{tabulary}
\usepackage{colortbl}
\usepackage{wrapfig}
\usepackage{amsmath, amssymb, amsfonts}
\usepackage{algorithm}
\usepackage{algpseudocodex}
\usepackage{makecell}

\definecolor{oursblue}{RGB}{239,246,255}

\definecolor{impgreen}{RGB}{34,139,34}
\definecolor{badred}{RGB}{190,70,70}

\newcommand{\ours}{\quad + \textsc{TrackCue}}

\newcommand{\gain}[1]{\textcolor{impgreen}{\scriptsize\,($\downarrow$#1)}}
\newcommand{\loss}[1]{\textcolor{badred}{\scriptsize\,($\uparrow$#1)}}

\newcommand{\bestgain}[2]{\makecell[c]{\textbf{#1}\\[-0.15em]\gain{#2}}}
\newcommand{\badloss}[2]{\makecell[c]{#1\\[-0.15em]\loss{#2}}}

\newcommand{\myparagraph}[1]{\vspace{2pt}\noindent{\bf #1}}

\definecolor{oursgray}{RGB}{235,235,235}
\definecolor{ourspurple}{RGB}{244,238,255} 

\title{Motion Cues from Image-based Point Tracking \\ for LiDAR Scene Flow Estimation}

\author{
  {\bfseries
  Youngdong Jang\textsuperscript{\normalfont *,1} \hspace{0.5em}
  Gyeongrok Oh\textsuperscript{\normalfont *,1} \hspace{0.5em}
  Jong Wook Kim\textsuperscript{\normalfont 1} \hspace{0.5em}
  Hyunju Ryu\textsuperscript{\normalfont 1}
  } \\[0.5em]
  {\bfseries
  Hyung-gun Chi\textsuperscript{\normalfont 2} \hspace{0.5em}
  SeungHyeon Kim\textsuperscript{\normalfont 1} \hspace{0.5em}
  Seungryong Kim\textsuperscript{\normalfont 3} \hspace{0.5em}
  Jonghyun Choi\textsuperscript{\normalfont 4} \hspace{0.5em}
  Sangpil Kim\textsuperscript{\normalfont $\dagger$,1}
  } \\
  \\
  \textsuperscript{\normalfont 1} Korea University \quad
  \textsuperscript{\normalfont 2} Purdue University \quad
  \textsuperscript{\normalfont 3} KAIST \quad 
  \textsuperscript{\normalfont 4} Hyundai Motor Company
}

\begin{document}

\maketitle
\begingroup
\renewcommand{\thefootnote}{}
\renewcommand{\footnoterule}{}
\footnotetext{\phantomsection\noindent* Equal contribution. \quad $\dagger$ Corresponding author.}
\endgroup

\begin{abstract}

Self-supervised LiDAR scene flow approaches rely on pseudo-labels that distinguish static and dynamic points, providing targeted supervision. These labels are typically derived from LiDAR geometry, making them unreliable in sparse or occluded regions. The resulting noise misguides motion supervision and degrades scene flow learning, yet existing methods remain largely dependent on the same geometric evidence.
To address this, we introduce \textsc{TrackCue}, a tracking-guided framework that transfers dense image trajectories into reliable static-dynamic supervision. \textsc{TrackCue} anchors tracking queries to projected LiDAR points, yielding image-space trajectories that provide motion evidence beyond sparse LiDAR geometry.
Since image-space displacement also reflects ego-vehicle motion, we present a visually consistent motion compensation strategy that compares each tracked trajectory with the rigid trajectory induced by ego-motion alone. It admits motion evidence only where both trajectories are valid and the deviation persists across frames, isolating true object motion while suppressing transient tracking errors. Finally, we perform visual motion cue lifting, which associates image-space object motion cues with LiDAR points for label refinement. As a result, \textsc{TrackCue} substantially improves the precision of dynamic pseudo-labels, providing more reliable supervision and leading to better self-supervised scene flow estimation.
\end{abstract}

\section{Introduction}


LiDAR scene flow estimation~\cite{li2021neural, li2023fast, li2025uniflowzeroshotlidarscene, zhang2026deltaflow} is essential for autonomous driving, as it provides point-level 3D motion from consecutive point clouds. However, accurate scene flow learning is challenging due to the high cost of ground-truth annotations. To reduce this reliance on manual labels, self-supervised approaches learn scene flow directly from point cloud sequences. 
Recent self-supervised approaches~\cite{vedder2024neural, hoffmann2025floxels, zhang2025himo, zhang2024seflow} can be divided into optimization-based and feed-forward methods. In this work, we focus on feed-forward methods~\cite{lin2025voteflow, zhang2025himo, vedder2024zeroflow}, which are more suitable for practical deployment than optimization-based methods by avoiding costly per-scene optimization at test time.


\begin{figure}[t!]
\begin{center}
\includegraphics[width=\linewidth]{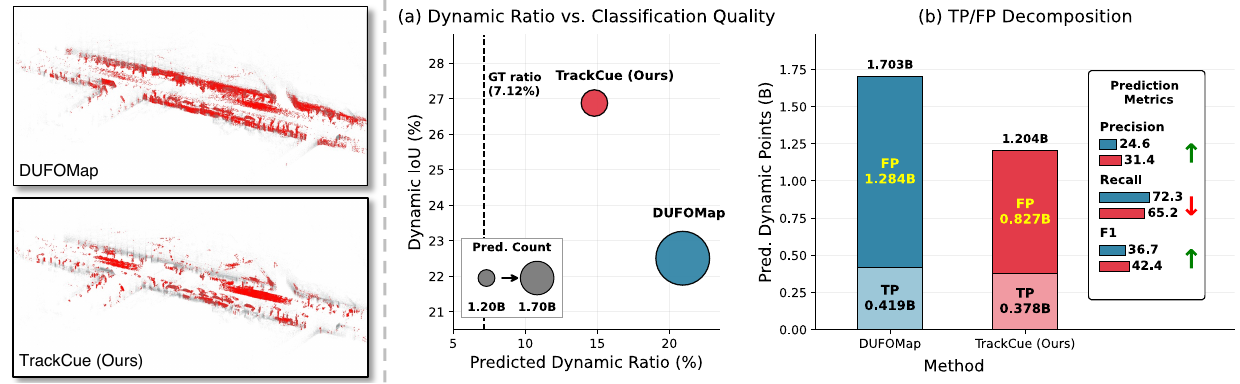}
\end{center}
\vspace{-1em}
\caption{Analysis of dynamic auto-label quality for self-supervised LiDAR scene flow.
(Left) \textsc{TrackCue} suppresses static regions while preserving true object motion.
(Right) Compared with DUFOMap~\cite{duberg2024dufomap}, \textsc{TrackCue} reduces dynamic over-prediction and false positives, improving Dynamic IoU, precision, and F1. The dashed line denotes the ground-truth dynamic ratio.
}
\label{fig:motive}
\vspace{-1em}
\end{figure}

A key challenge in this setting is the severe imbalance between static and dynamic points. Most LiDAR points in driving scenes are static. Without accounting for this imbalance, self-supervision biases the model toward predicting zero flow. To address this, existing approaches~\cite{zhang2024seflow, zhang2025himo} first separate static and dynamic points, applying different supervision to each. 
This static-dynamic separation typically relies on ray-casting-based dynamic-awareness mapping, such as DUFOMap~\cite{duberg2024dufomap}, which is constructed only from LiDAR point clouds. Specifically, ray-casting identifies regions traversed by rays as voids and classifies points later observed in those regions as dynamic. 
However, this procedure depends on dense and consistent LiDAR observations. 
Sparse or occluded observations yield unreliable void estimates, causing static points to be mislabeled as dynamic and moving points to be missed.
The resulting noise misguides motion supervision and degrades scene flow learning, while refinements based solely on LiDAR geometry~\cite{zhang2025himo} remain vulnerable to sparsity and occlusion.

A natural approach is to track points in RGB images, where object motion remains visually continuous despite sparse or occluded LiDAR. However, naive point tracking is insufficient. Image motion mixes object and ego-induced motion, and tracking suffers from transient correspondence and visibility errors. 
Even with reliable trajectories, image-space motion cues do not directly transfer to sparse LiDAR observations.

To address these issues, we introduce \textsc{TrackCue}, a tracking-guided framework that transfers temporally consistent image trajectories into reliable static-dynamic supervision. \textsc{TrackCue} anchors tracking queries to projected LiDAR points, maintaining correspondence between image-space trajectories and 3D observations. 
As shown in Figure~\ref{fig:figure3}, static points exhibit substantial image-space displacement under ego-vehicle motion. Since this displacement alone cannot distinguish object motion from ego-motion, we present a visually consistent motion compensation strategy. Specifically, \textsc{TrackCue} resolves this ambiguity by comparing each tracked trajectory with the rigid trajectory of the same point under ego-motion alone, and identifies points that consistently deviate from it as independently moving.
A joint visibility mask with temporal voting then restricts this comparison to frames where both the tracked trajectory and its rigid projection are valid, and requires the deviation to persist across frames, suppressing transient tracking errors. Finally, \textsc{TrackCue} performs visual motion cue lifting, transferring the ego-compensated cues back to the LiDAR domain by associating their image-space locations with projected LiDAR points through nearest-neighbor search. The resulting labels complement LiDAR-only observations and provide more reliable supervision for self-supervised scene flow learning.

Figure~\ref{fig:motive} shows that \textsc{TrackCue} produces more reliable pseudo-labels than DUFOMap~\cite{duberg2024dufomap}. It reduces false-positive dynamic labels by 35.6\%, while improving precision from 24.6\% to 31.4\% and F1 score from 36.7\% to 42.4\%. Our results show that \textsc{TrackCue} improves
dynamic label reliability and scene flow performance. Our contributions are summarized as follows:



\begin{itemize} [leftmargin=*]
    \item We introduce \textsc{TrackCue}, a tracking-guided framework that leverages dense image trajectories to complement sparse LiDAR geometry for reliable static-dynamic supervision.
    \item We propose a visually consistent motion compensation strategy with joint visibility masking and temporal voting to distinguish object-motion cues from ego-induced apparent motion while suppressing transient tracking errors.
    \item We present visual motion cue lifting, which transfers reliable image-space motion evidence back to the LiDAR domain for static-dynamic label refinement, leading to improved pseudo-label quality and downstream scene flow performance.
\end{itemize}

\section{Related Works}
\noindent \textbf{Scene Flow Estimation} aims to estimate the 3D motion vector of moving objects in dynamic scenes. In autonomous driving, existing works have explored both supervised~\cite{li2025uniflowzeroshotlidarscene, zhang2026deltaflow, jund2021scalable, khatri2024can, khoche2025ssf, kim2025flow4d} and self-supervised learning~\cite{zhang2026teflow, zhang2025himo, zhang2024seflow, lin2025voteflow, vedder2024zeroflow, vedder2024neural, hoffmann2025floxels, lin2024icp, li2021neural, li2023fast} paradigms. Recently, self-supervised approaches have gained increasing attention, as dense 3D annotations are costly and difficult to scale. One prominent line of work~\cite{vedder2024neural, hoffmann2025floxels, li2021neural, li2023fast} performs per-scene optimization at test time. While these approaches exploit long-term temporal context to estimate accurate flow fields, their computational overhead limits real-time deployment. Another line of work~\cite{zhang2026teflow, zhang2025himo, zhang2024seflow, lin2025voteflow, vedder2024zeroflow} uses feed-forward models pretrained on large-scale unlabeled data, enabling flow prediction in a single forward pass. Importantly, the self-supervised approaches highly depend on pseudo-labels based on dynamic object clustering. 
DUFOMap~\cite{duberg2024dufomap} is an efficient dynamic-awareness mapping framework that separates moving objects from the static environment. It enables reliable flow supervision across consecutive point clouds, even without manually annotated labels. 
However, inherent limitations of LiDAR, including sparse observations, temporally varying point measurements, and occlusions, can lead to dynamic object misclassifications.
Such erroneous dynamic labels disrupt training stability and ultimately degrade the model performance.
To this end, our proposed \textsc{TrackCue} refines dynamic object clustering informed by dense visual cues from RGB images.

\noindent \textbf{Multi-modal Motion Understanding} has become a widely adopted paradigm in safety-critical vision systems, particularly in autonomous driving perception. By combining the accurate geometric localization of LiDAR with the dense appearance and semantic cues of RGB images, camera--LiDAR fusion~\cite{liu2023bevfusion, zhang2026bevdilation, wang2023unitr} has shown strong effectiveness in various 3D perception tasks, including object detection~\cite{huang2024detecting, chang2024cmda}, semantic segmentation~\cite{zheng2024learning, zhang2023delivering}, and occupancy prediction~\cite{pan2024co, wang2023openoccupancy}. Furthermore, cross-modal distillation~\cite{sautier2022image, Xu_2025_ICCV, liu2023segment} has also been explored to transfer complementary knowledge across modalities, thereby improving feature representation for downstream tasks. 
Despite these advances in 3D perception, multi-modal reasoning remains relatively underexplored in LiDAR scene flow estimation~\cite{liu2022camliflow, li2025uniflowzeroshotlidarscene}, especially under self-supervised settings. Inspired by
these developments, our work pioneers a new perspective on scene flow learning, which improves the reliability of self-supervised dynamic labels.

\noindent \textbf{Point Tracker} in videos has made tremendous progress in recent years. It demonstrates a remarkable ability to understand the temporal evolution of scenes while capturing the physical dynamics of moving instances. Originally, the task of tracking any point~\cite{harley2022particle, doersch2022tap} was first formalized by the classical optical flow estimation problem, consistently generating point trajectories with extended temporal window instead of 2-frames. Afterwards, the series of TAP~\cite{doersch2022tap, doersch2023tapir, vecerik2023robotap, doersch2024bootstap, koppula2024tapvid, zholus2025tapnext} and CoTracker~\cite{karaev2024cotracker, karaev2025cotracker3} are built upon modern transformer architectures and trained on large-scale video datasets, including both labeled and unlabeled, real and synthetic data. Beyond these representative works, a growing number of studies~\cite{harley2025alltracker, cho2024local} have further advanced this direction, exploring more robust and scalable point tracking frameworks and broadening their applicability~\cite{wang2024vggsfm, le2024dense, huang2025segment, rajivc2025segment}. In this paper, we argue that foundation point trackers can serve as a powerful source of visual motion cues for LiDAR scene flow estimation, complementing sparse geometric observations with dense and temporally consistent RGB-based trajectories.

\section{Method}

\subsection{Preliminary}
\paragraph{Problem Statement.}
LiDAR scene flow estimation aims to predict a 3D motion vector $\mathbf{f}_{i,t} \in \mathbb{R}^3$ for each LiDAR point, describing its motion to the next frame. 
Given consecutive LiDAR point clouds $\{\mathcal{P}_t, \mathcal{P}_{t+1}, \dots \}$, $\mathcal{P}_t=\{\mathbf{p}_{i,t}\}_{i=1}^{M_t}$ denotes the point cloud at time $t$ with $M_t$ points. 
For all points in $\mathcal{P}_t$, the predicted scene flow is represented as $\mathcal{F}_t = \{\mathbf{f}_{i,t}\}_{i=1}^{M_t}$ from $\mathcal{P}_t$ to $\mathcal{P}_{t+1}$.
Since autonomous driving scenes are observed from a moving vehicle, the scene flow is decomposed into ego-motion flow $\mathcal{F}^{\mathrm{ego}}_t$ and residual flow $\Delta \mathcal{F}_t$ as follows:
\begin{equation}
\mathcal{F}_t = \mathcal{F}^{\mathrm{ego}}_t + \Delta \mathcal{F}_t.
\label{eq:flow_decomp}
\end{equation}
While $\mathcal{F}^{\text{ego}}$ can be derived from vehicle odometry, the network is trained to estimate $\Delta \mathcal{F}_t$.

\paragraph{Static-Dynamic Self-Supervision.}
To derive supervision without labeled data, self-supervised approaches first divide points into static and dynamic sets, denoted as $\mathcal{P}^{\mathrm{s}}_t$ and $\mathcal{P}^{\mathrm{d}}_t$, respectively. 
This classification allows different supervision signals to be applied according to the motion type of each point. The overall self-supervised objective is formulated by combining static and dynamic losses:
\begin{equation}
\mathcal{L}_{\mathrm{ssl}} 
= \mathcal{L}_{\mathrm{s}}(\mathcal{P}^{\mathrm{s}}_t)
+ \mathcal{L}_{\mathrm{d}}(\mathcal{P}^{\mathrm{d}}_t),
\label{eq:ssl_loss}
\end{equation}
where $\mathcal{L}_{s}$ constrains static points to have near-zero residual motion, whereas $\mathcal{L}_{d}$ supervises dynamic points through Chamfer distance and dynamic clustering terms. Therefore, reliable static-dynamic classification is crucial, as it determines whether each point receives appropriate supervision. More detailed descriptions of the labeling process are provided in Appendix~\ref{app:b}.

\begin{figure}[t!]
\begin{center}
\includegraphics[width=\linewidth]{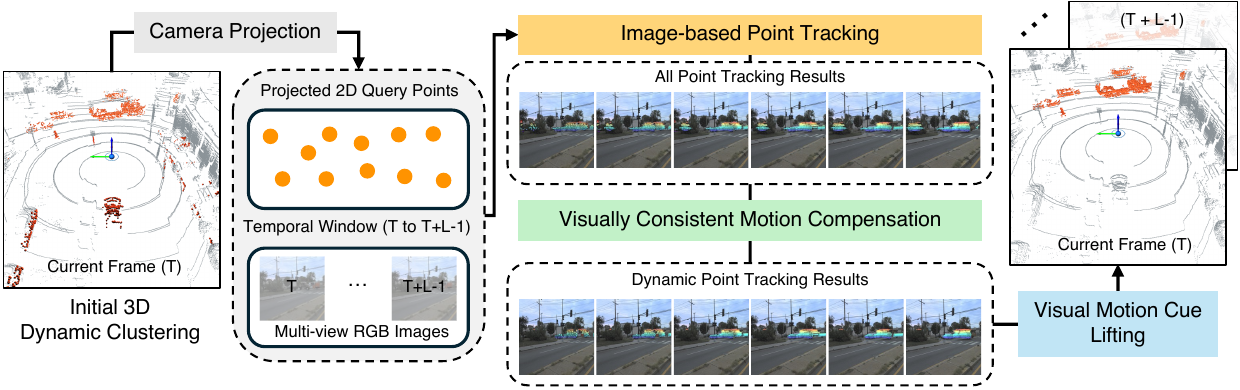}
\end{center}
\vspace{-1.em}
\caption{{Overview of the \textsc{TrackCue} framework.}}
\label{fig:figure2}
\end{figure}

\subsection{Overview}
\textsc{TrackCue} improves static-dynamic labeling by transferring motion evidence from RGB images into the LiDAR domain. As shown in Figure~\ref{fig:figure2}, \textsc{TrackCue} tracks projected LiDAR points in RGB images, compensates for ego-induced motion, suppresses unreliable trajectories, and lifts the resulting motion cues back to the LiDAR domain. The pipeline consists of three main components:
\begin{itemize} [leftmargin=*]
    \item \textbf{Image-based Point Tracking:} Anchors queries to projected LiDAR points and extracts temporally consistent trajectories from multi-frame images.
    These trajectories provide temporally consistent visual motion cues for identifying dynamic points that are ambiguous from LiDAR alone.
    \item \textbf{Visually Consistent Motion Compensation:} Resolves motion ambiguity and trajectory uncertainty. It constructs the rigid trajectory of each query under ego-motion alone and measures the residual against the observed track. A joint visibility mask removes invalid frames, while temporal voting retains only deviations that persist across frames.
    \item \textbf{Visual Motion Cue Lifting:} Bridges the image-to-LiDAR correspondence gap. It lifts ego-compensated image motion cues to LiDAR points through nearest-neighbor search, refining static-dynamic labels with image-space motion evidence.
\end{itemize}

\subsection{Image-based Point Tracking}
We propose an Image-based Point Tracking module to obtain temporally consistent motion cues from RGB images. While prior self-supervised methods derive static-dynamic labels from LiDAR geometry alone, our module exploits temporally continuous visual cues to complement sparse or occluded observations.

\paragraph{LiDAR-to-Image Projection.}
To initialize the image-based tracking, we first project the LiDAR points labeled by DUFOMap~\citep{duberg2024dufomap} onto the image plane. For each LiDAR point $\mathbf{p}_{i,t} \in \mathcal{P}^{\mathrm{dufo}}_t$, its pixel coordinate $\mathbf{u}_{i,t}=(u_{i,t}, v_{i,t})^\top$ is derived by applying the extrinsic transformation $\textbf{T}$ and the camera intrinsic matrix $\textbf{K}$:
\begin{equation}
    \begin{bmatrix}\mathbf{p}_{i,t}^{\mathrm{c}} \\ 1\end{bmatrix} = \mathbf{T}  \cdot
    \begin{bmatrix} \mathbf{p}_{i,t} \\ 1 \end{bmatrix} 
     , 
     \quad
    \mathbf{u}_{i,t}= \pi\left(\mathbf{K} \cdot \mathbf{p}_{i,t}^{\mathrm{c}}\right) 
    , 
\end{equation}
where $\mathbf{p}_{i,t}^{\mathrm{c}} = (x_{i,t}^{\mathrm{c}}, y_{i,t}^{\mathrm{c}}, z_{i,t}^{\mathrm{c}})^\top$ and $\pi(\cdot)$ denote the camera coordinate and the standard perspective projection, respectively. To ensure reliable tracking initialization, we retain only projected points that satisfy the following constraints:
\begin{equation}
z_{i,t}^{\mathrm{c}} > d_{\mathrm{min}}, \quad b \le u_{i,t} < W - b, \quad b \le v_{i,t} < H - b,
\label{eq:project_constraints}
\end{equation}
where $z_{i,t}^{c}$, $(W, H)$, and $b$ denote the camera-frame depth, resolution, and boundary margin, respectively. The retained projections form the query set $\mathcal{Q}_t = \{ \mathbf{q}_{i,t} \}_{i=1}^{N_t}$.

\paragraph{Point Tracking.}
Given $\mathcal{Q}_t$, we estimate trajectories over a clip of length $L$ using a pre-trained point tracker~\citep{karaev2025cotracker3, harley2025alltracker}. Unlike conventional optical flow, point trackers maintain long-term temporal consistency and remain robust under occlusion. For each initial query $\mathbf{q}_{i,t}$, the point tracker predicts its corresponding image position in each subsequent frame. We denote the tracked trajectory as:
\begin{equation}
    \hat{\mathcal{T}}_i = \left\{\hat{\mathbf{u}}_{i,k}\right\}_{k=t}^{t+L-1},
\end{equation}
where $\hat{\mathbf{u}}_{i,k}$ is the estimated pixel coordinate at frame $k$.
The tracker also predicts a visibility mask $w^{\mathrm{tracker}}_{i,k} \in \{0,1\}$, where $w^{\mathrm{tracker}}_{i,k}=1$ indicates that the tracked point is predicted to be visible at frame \(k\). Since tracking operates in the image domain, $\hat{\mathcal{T}}_i$ can remain continuous across frames even when the corresponding LiDAR observations are sparse or absent.


\begin{wrapfigure}{r}{0.50\textwidth}
    \vspace{-1.5em}
    \centering
    \includegraphics[width=0.46\textwidth]{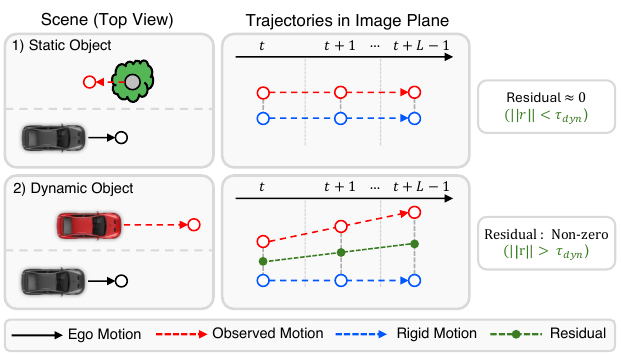}
    \vspace{-0.6em}
    \caption{{Illustration of our motion compensation.}}
    \label{fig:figure3}
    \vspace{-1.2em}
\end{wrapfigure}

\subsection{Visually Consistent Motion Compensation}
We present a visually consistent motion compensation to distinguish true object motion from ego-induced apparent motion in image-space trajectories. 
Although a point tracker provides robust trajectories, its predictions capture image-space motion, which contains both independent object motion and ego-induced apparent motion. Thus, motion compensation is required to separate true object motion from ego-induced apparent motion in the image-space trajectories. To ensure that the motion evidence is reliable, a joint visibility mask filters out frames where either trajectory is invalid, while temporal voting retains only deviations that persist across multiple frames.

\paragraph{Rigid Trajectory and Residual.}
Assuming each query $\mathbf{q}_{i,t}$ corresponds to a LiDAR point in the static scene, we construct a rigid trajectory $\hat{\mathcal{T}}_i^{\mathrm{rigid}} = \{ \hat{\mathbf{u}}_{i,k}^{\mathrm{rigid}} \}_{k=t}^{t+L-1}$ for that point. This trajectory represents image-space displacement induced solely by ego-motion. 
Using the global ego-poses $\mathbf{G} \in SE(3)$, the rigid pixel coordinate $\hat{\mathbf{u}}_{i,k}^{\mathrm{rigid}}$ at frame $k$ is obtained by applying the relative ego-motion from frame $t$ to $k$ and projecting the transformed point onto the target image plane:
\begin{equation}
\hat{\mathbf{u}}_{i,k}^{\mathrm{rigid}} = \pi \left( \mathbf{K} \cdot \mathbf{T} \cdot \mathbf{G}_k^{-1} \cdot \mathbf{G}_t \cdot \begin{bmatrix} \mathbf{p}_{i,t} \\ 1 \end{bmatrix} \right), 
\end{equation}
where $\mathbf{G}_t$ and $\mathbf{G}_k$ represent the global ego-poses at the reference and target frames, respectively. The term $\mathbf{G}_k^{-1} \cdot \mathbf{G}_t$ gives the relative transformation from frame t to frame $k$. To ensure a valid rigid projection, we construct a rigid visibility mask $w_{i,k}^{\mathrm{rigid}} \in \{0, 1\}$ for each $\hat{\mathbf{u}}_{i,k}^{\mathrm{rigid}}$, using the projection constraints in Eq.~\ref{eq:project_constraints}. The residual between the tracked and rigid trajectories isolates motion unexplained by ego-motion:
\begin{equation}
\label{eq:residual}
    r_{i,k} = \parallel \hat{\mathbf{u}}_{i,k} - \hat{\mathbf{u}}_{i,k}^{\mathrm{rigid}} \parallel_2.
\end{equation}
For static points, the residual $r_{i,k}$ remains close to zero, whereas true dynamic points exhibit non-zero residuals. Therefore, we use the residual as the criterion for identifying independently moving points.

\paragraph{Joint Visibility and Temporal Voting.}
The residual is reliable only when both the tracked and rigid trajectories are valid. If the tracked point is occluded or the rigid projection falls outside the image, invalid coordinates can produce a large residual unrelated to object motion. Such a residual may be interpreted as movement. We therefore define a joint visibility mask $w_{i,k}^{\mathrm{joint}}$ that admits a frame into the comparison only when both are valid:
\begin{equation}
    w_{i,k}^{\mathrm{joint}} = w_{i,k}^{\mathrm{tracker}} \cdot w_{i,k}^{\mathrm{rigid}}.
\end{equation}
Even under valid visibility, a large residual at a single time step can arise from transient correspondence errors. To suppress such errors, we apply temporal voting by counting the jointly valid frames in which the residual exceeds $\tau_{\mathrm{dyn}}$. Based on $V_i$, we collect the candidate moving trajectory indices:
\begin{equation}
\label{eq:mask}
    V_i = \sum_{k=t}^{t+L-1} \mathbf{1} \left[r_{i,k} > \tau_{\mathrm{dyn}}\right] w_{i,k}^{\mathrm{joint}}, \quad     \mathcal{I}^{\mathrm{mov}} = \left\{i \mid V_i \geq N_{\mathrm{move}} \right\},
\end{equation}
where $\mathbf{1}[\cdot]$, $\tau_{\mathrm{dyn}}$, and $N_{\mathrm{move}}$ denote the indicator function, residual threshold and minimum number of moving frames, respectively. For each camera view, we retain the selected trajectories only when at least \(N_{\mathrm{point}}\) moving queries are found, requiring the motion cue to be supported by multiple queries rather than a few isolated detections:
\begin{equation}
\mathcal{T}^{\mathrm{mov}}
=
\begin{cases}
\left\{
\hat{\mathcal{T}}_i
\mid i \in \mathcal{I}^{\mathrm{mov}}
\right\},
& |\mathcal{I}^{\mathrm{mov}}| \geq N_{\mathrm{point}}, \\
\emptyset,
& \text{otherwise}.
\end{cases}
\end{equation}
Although $\mathcal{T}^{\mathrm{mov}}$ provides reliable evidence of independent motion, it remains defined on image trajectories and is not yet assigned to LiDAR points.

\subsection{Visual Motion Cue Lifting}
We lift the selected moving trajectories $\mathcal{T}^{\mathrm{mov}}$ to the LiDAR domain to obtain point-level dynamic evidence. Since these trajectories originate from only a subset of LiDAR points, they do not provide one-to-one correspondence with all dynamic points. Thus, we propagate the visual motion evidence to nearby LiDAR points. For each frame $k$, we use the visible positions of the selected trajectories and project the LiDAR points in the same frame onto the image plane. A KD-tree $\mathcal{D}_k$ is then constructed over the valid projected LiDAR coordinates. Given a selected trajectory $\hat{\mathcal{T}}_i \in \mathcal{T}^{\mathrm{mov}}$, its visible image-space position $\hat{\mathbf{u}}_{i,k}$ is associated with the $K_{\mathrm{nn}}$ nearest projected LiDAR points:
\begin{equation}
\mathcal{N}_{i,k} = \mathcal{D}_k\!\left(\hat{\mathbf{u}}_{i,k},\, K_{\mathrm{nn}}\right),
\qquad
d_{j,k} = \lVert \hat{\mathbf{u}}_{i,k} - \mathbf{u}_{j,k} \rVert_2 ,
\end{equation}
where $\mathcal{N}_{i,k}$ denotes the index set of the $K_{\mathrm{nn}}$ nearest valid projected LiDAR points in frame $k$. Each matched LiDAR point is assigned a dynamic cue only when the nearest-neighbor distance is smaller than a lifting radius  $\tau_{\mathrm{lift}}$:
\begin{equation}
e_{j,k} = \mathbf{1}\!\left[\, j \in \mathcal{N}_{i,k}
\;\wedge\; d_{j,k} < \tau_{\mathrm{lift}} \,\right].
\end{equation}
Here, $e_{j,k}$ indicates whether LiDAR point $\mathbf{p}_{j,k}$ receives a lifted visual motion cue. These lifted cues complement the DUFOMap-based labels by recovering dynamic points that are difficult to identify from sparse LiDAR geometry alone. The resulting labels refine $\mathcal{P}^{\mathrm{dufo}}$ into $\mathcal{P}^{\textsc{TrackCue}}$, providing more reliable static-dynamic supervision.

\section{Experiments}

\subsection{Experimental Setup}
\myparagraph{Datasets.}
We evaluate on Argoverse 2~\cite{Argoverse2}, which contains 700 training and 150 validation sequences. The official public challenge~\cite{li2025uniflowzeroshotlidarscene} further evaluates methods on a diverse test set comprising five widely used LiDAR scene flow datasets: Argoverse 2, Waymo~\cite{sun2020scalability}, nuScenes~\cite{caesar2020nuscenes}, TruckScenes~\cite{fent2024man}, and AEVAScenes~\cite{aevascenes}. The test set includes 458 scenes and 9,613 frames randomly sampled from five source datasets. It provides broad coverage across driving environments and sensor configurations, such as urban and highway scenes captured with 32- and 64-beam LiDARs. For leaderboard evaluation, all scene flow models are trained only on the Argoverse 2 training split, with no additional scene flow training data.

\myparagraph{Metrics.}
We report two representative scene flow metrics: three-way End Point Error~(EPE) and Dynamic Bucket-Normalized EPE~\cite{khatri2024can}. EPE measures the $\ell_2$ distance between predicted and ground-truth flow vectors in centimeters. Three-way EPE averages EPE over Foreground Dynamic~(FD), Foreground Static~(FS), and Background Static~(BS) points, while Bucket-Normalized EPE normalizes errors across speed buckets and semantic classes. 

\myparagraph{Baselines.}
We validate \textsc{TrackCue} by applying it to three self-supervised scene flow frameworks: SeFlow~\cite{zhang2024seflow}, SeFlow++~\cite{zhang2025himo}, and TeFlow~\cite{zhang2026teflow}. SeFlow introduces self-supervised learning signals derived from dynamic point classification. SeFlow++ further leverages nearest-neighbor correspondences to filter static points according to their matching distances, improving geometric temporal consistency. TeFlow improves supervision reliability by integrating past trajectories within a multi-frame framework, while still relying on DUFOMap~\cite{duberg2024dufomap}-based static-dynamic classification. For a fair comparison, we build on the official implementations of all baselines and retain their provided training configurations. We provide further experimental details in Appendix~\ref{app:a}.

\begin{table*}[t!]
\centering
\small
\setlength{\tabcolsep}{0.36em}
\renewcommand{\arraystretch}{1.08}
\caption{
Experimental results on the Argoverse 2 \underline{leaderboard}
~\cite{li2025uniflowzeroshotlidarscene}.
We report Dynamic Bucket Normalized EPE~\cite{khatri2024can}, where lower is better. Small arrow indicates the scaled absolute score difference
from the corresponding baseline, i.e.,
$|\mathrm{Ours}-\mathrm{Baseline}|\times100$. 
\textcolor{impgreen}{Green} downward arrows denote improvement, while \textcolor{badred}{red} upward arrows denote degradation.
}
\label{tab:av2_test}
\resizebox{\linewidth}{!}{
\begin{tabular}{
l
cc
!{\color{black!18}\vrule width 0.4pt}
cc
!{\color{black!18}\vrule width 0.4pt}
cc
!{\color{black!18}\vrule width 0.4pt}
cc
!{\color{black!18}\vrule width 0.4pt}
cc
!{\color{black!18}\vrule width 0.4pt}
cc
}
\toprule
\multirow{2.5}{*}{Method}
& \multicolumn{2}{c}{Overall}
& \multicolumn{2}{c}{AV2~\cite{Argoverse2}}
& \multicolumn{2}{c}{Waymo~\cite{sun2020scalability}}
& \multicolumn{2}{c}{TruckScenes~\cite{fent2024man}}
& \multicolumn{2}{c}{NuScenes~\cite{caesar2020nuscenes}}
& \multicolumn{2}{c}{Aeva~\cite{aevascenes}} \\
\cmidrule(lr){2-3}
\cmidrule(lr){4-5}
\cmidrule(lr){6-7}
\cmidrule(lr){8-9}
\cmidrule(lr){10-11}
\cmidrule(lr){12-13}
& Mean & 0--35m
& 0--35m & 35--70m
& 0--35m & 35--70m
& 0--35m & 35--70m
& 0--35m & 35--70m
& 0--35m & 35--70m \\
\midrule

SeFlow~\citep{zhang2024seflow}
& 0.7367 & 0.6289
& 0.3113 & 0.6638
& 0.3494 & 0.7639
& 0.8885 & 0.8509
& 0.7543 & 1.0514
& 0.8412 & 0.8926 \\

\rowcolor{oursblue}
\ours
& \bestgain{0.7344}{0.23} & \bestgain{0.6286}{0.03}
& \bestgain{0.3070}{0.43} & \badloss{0.6958}{3.20}
& \bestgain{0.3234}{2.60} & \bestgain{0.7479}{1.60}
& \badloss{0.8968}{0.83} & \badloss{0.8675}{1.66}
& \badloss{0.7933}{3.90} & \bestgain{0.9910}{6.04}
& \bestgain{0.8227}{1.85} & \badloss{0.8990}{0.64} \\

\addlinespace[0.25em]
\midrule

SeFlow++~\citep{zhang2025himo}
& 0.7025 & 0.5919
& 0.2827 & 0.6552
& 0.3192 & 0.7253
& \textbf{0.8555} & 0.8473
& 0.6965 & 0.9748
& 0.8057 & 0.8631 \\

\rowcolor{oursblue}
\ours
& \bestgain{0.6619}{4.06}
& \bestgain{0.5537}{3.82}
& \bestgain{0.2781}{0.46}
& \bestgain{0.6313}{2.39}
& \bestgain{0.3041}{1.51}
& \bestgain{0.6339}{9.14}
& \badloss{0.8565}{0.10}
& \bestgain{0.7978}{4.95}
& \bestgain{0.6550}{4.15}
& \bestgain{0.9632}{1.16}
& \bestgain{0.6748}{13.09}
& \bestgain{0.8238}{3.93} \\

\addlinespace[0.25em]
\midrule

TeFlow~\citep{zhang2026teflow}
& 0.6125 & 0.3116
& 0.2019 & 0.8826
& 0.1979 & 0.8920
& 0.3884 & \textbf{0.9108}
& 0.3822 & \textbf{0.9848}
& 0.3875 & \textbf{0.8969} \\

\rowcolor{oursblue}
\ours
& \bestgain{0.6057}{0.68}
& \bestgain{0.2987}{1.29}
& \bestgain{0.1966}{0.53}
& \bestgain{0.8734}{0.92}
& \bestgain{0.1870}{1.09}
& \bestgain{0.8872}{0.48}
& \bestgain{0.3762}{1.22}
& \badloss{0.9123}{0.15}
& \bestgain{0.3721}{1.01}
& \badloss{0.9867}{0.19}
& \bestgain{0.3619}{2.56}
& \badloss{0.9026}{0.57} \\

\bottomrule
\end{tabular}
}
\vspace{-2em}
\end{table*}

\begin{table*}[t!]
\centering
\footnotesize
\setlength{\tabcolsep}{0.55em}
\renewcommand{\arraystretch}{1.12}
\caption{
Experimental results on the Argoverse 2 \underline{validation set}
~\cite{Argoverse2}. Small arrow indicates the absolute score difference from corresponding baseline.
\textcolor{impgreen}{Green} downward arrows denote improvement, while \textcolor{badred}{red} upward arrows denote degradation.
}
\label{tab:av2_val}
\begin{tabular}{l c cccc ccccc}
\toprule
\multirow{2.5}{*}{Method}
& \multirow{2.5}{*}{\#Frames}
& \multicolumn{4}{c}{Three-way EPE (cm) $\downarrow$}
& \multicolumn{5}{c}{Dynamic Bucket-Normalized $\downarrow$} \\
\cmidrule(lr){3-6} \cmidrule(lr){7-11}
&
& Mean & FD & FS & BS
& Mean & CAR & OTHER & PED. & VRU \\


\midrule
\multicolumn{11}{l}{\textit{Optimization-based}} \\
\midrule

NSFP~\citep{li2021neural}
& 2
& 7.19 & 12.44 & 4.52 & 4.62
& 0.406 & 0.218 & 0.284 & 0.698 & 0.422 \\

FastNSF~\citep{li2023fast}
& 2
& 8.32 & 13.09 & 5.73 & 6.15
& 0.365 & 0.236 & 0.405 & 0.501 & 0.318 \\

ICP-Flow~\citep{lin2024icp}
& 2
& 6.44 & 12.40 & 3.75 & 3.19
& 0.346 & 0.229 & 0.354 & 0.450 & 0.351 \\

\midrule
\multicolumn{11}{l}{\textit{Feed-forward}} \\
\midrule

ZeroFlow~\citep{vedder2024zeroflow}
& 2
& 6.60 & 12.90 & 3.72 & 3.16
& 0.565 & 0.275 & 0.435 & 0.894 & 0.658 \\

VoteFlow~\citep{lin2025voteflow}
& 2
& 5.56 & 13.11 & 2.12 & 1.44
& 0.335 & 0.223 & 0.342 & 0.429 & 0.347 \\

\addlinespace[0.15em]
SeFlow~\citep{zhang2024seflow}
& 2
& 4.88 & 11.37 & 1.97 & 1.31
& 0.372 & 0.215 & 0.378 & 0.551 & 0.343 \\

\rowcolor{oursblue}
\ours
& 2
& \bestgain{4.56}{0.32}
& \bestgain{10.58}{0.79}
& \bestgain{1.96}{0.01}
& \bestgain{1.14}{0.17}
& \bestgain{0.352}{0.020}
& \bestgain{0.194}{0.021}
& \bestgain{0.375}{0.003}
& \bestgain{0.518}{0.033}
& \bestgain{0.320}{0.023} \\

\addlinespace[0.25em]
SeFlow++~\citep{zhang2025himo}
& 3
& 4.78 & 9.48 & 3.05 & 1.80
& 0.328 & 0.182 & 0.334 & 0.508 & 0.290 \\

\rowcolor{oursblue}
\ours
& 3
& \bestgain{4.29}{0.49}
& \bestgain{9.15}{0.33}
& \bestgain{2.39}{0.66}
& \bestgain{1.31}{0.49}
& \bestgain{0.298}{0.030}
& \bestgain{0.176}{0.006}
& \bestgain{0.320}{0.014}
& \bestgain{0.422}{0.086}
& \bestgain{0.276}{0.014} \\

\addlinespace[0.25em]
TeFlow~\citep{zhang2026teflow}
& 5
& 3.59 & 7.18 & \textbf{1.36} & 2.23
& 0.222 & \textbf{0.140} & 0.285 & \textbf{0.284} & \textbf{0.179} \\

\rowcolor{oursblue}
\ours
& 5
& \bestgain{3.50}{0.09}
& \bestgain{7.01}{0.17}
& \badloss{1.51}{0.15}
& \bestgain{1.99}{0.24}
& \bestgain{0.218}{0.004}
& \badloss{0.145}{0.005}
& \bestgain{0.234}{0.051}
& \badloss{0.302}{0.018}
& \badloss{0.193}{0.014} \\

\bottomrule
\end{tabular}
\vspace{-1em}
\end{table*}

\subsection{LiDAR Scene Flow Estimation Results}
\myparagraph{Quantitative Analysis.}
Table~\ref{tab:av2_test} reports scene flow estimation results on the Argoverse 2 challenge across five datasets. 
As seen from the table, \textsc{TrackCue} improves flow predictions across most datasets and distance ranges. The larger improvement for SeFlow++ highlights the importance of reliable static and dynamic separation for self-supervision. While SeFlow and TeFlow rely on DUFOMap-based pseudo-labels, SeFlow++ reassigns the dynamic labels by applying geometric distance-based filtering. Note that \textsc{TrackCue} further alleviates the sparsity and ambiguity inherent to LiDAR observations through temporally continuous motion cues from RGB images, ultimately leading to larger performance gains. 
\textsc{TrackCue} also yields substantial improvements across diverse datasets under the zero-shot evaluation. For example, \textsc{TrackCue} achieves a 13.09 point reduction in Dynamic Bucket-Normalized EPE on AEVAScenes, despite its different sensor configuration.
These observations suggest that high-fidelity pseudo-labels improve the cross-dataset generalization of scene flow models. Detailed leaderboard results are provided in Table~\ref{tab:av2_test_av2} in Appendix~\ref{app:c}.
Table~\ref{tab:av2_val} presents results on the Argoverse 2 validation split. \textsc{TrackCue} shows the effectiveness of deriving visual motion supervision for scene flow prediction through detailed class-wise performance.
The key insight is that \textsc{TrackCue} consistently reduces Foreground Dynamic (FD) error across all three baselines, supporting the effectiveness of the refined labels for dynamic-object supervision.


\begin{figure}[t!]
\begin{center}
\includegraphics[width=\linewidth]{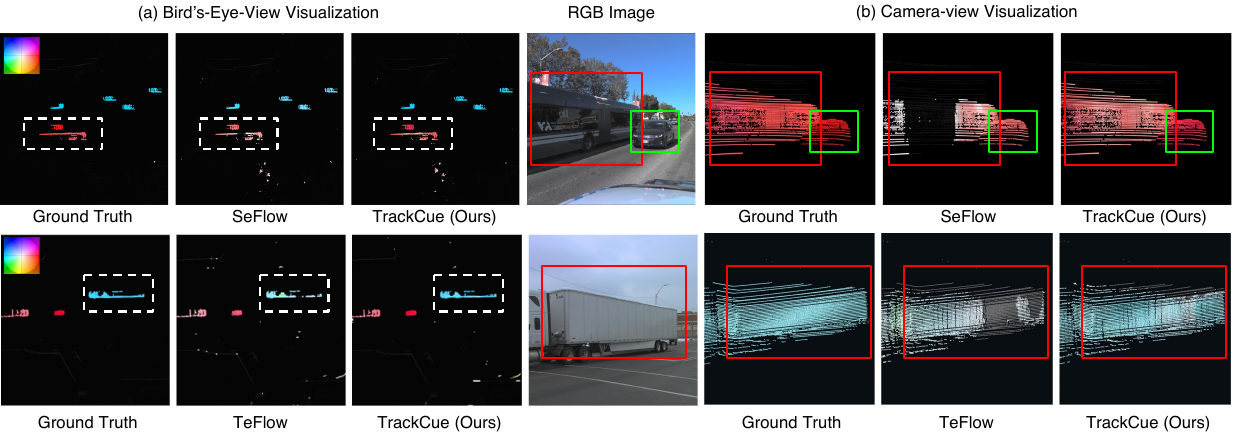}
\end{center}
\vspace{-1em}
\caption{We visualize the flow estimation results in terms of two complementary perspectives. Motion is color-coded, where hue represents direction and intensity indicates speed.}
\label{fig:comp_flow}
\end{figure}

\myparagraph{Qualitative Analysis.}
Figure~\ref{fig:comp_flow} visualizes the flow estimation results on some challenging scenarios. Here, the top example shows that \textsc{TrackCue} accurately estimates flow velocity, as indicated by the vivid color-coded motion. \textsc{TrackCue} also filters out points that are incorrectly classified as dynamic, preventing erroneous flow estimation in the background.  
In contrast, the baselines fail to predict the correct magnitude of the flow vectors, while also assigning non-zero motion to static background regions.
Furthermore, the bottom example highlights the superiority of dense RGB motion cues.
Specifically, the baselines miss motion predictions for parts of large instances. 
Since the central regions of these objects remain non-empty over short temporal windows, corresponding points are consequently misclassified as static by DUFOMap.
However, \textsc{TrackCue} consistently assigns dynamic labels by focusing on image-space trajectories, rather than relying solely on sparse LiDAR observations. We present further qualitative results of \textsc{TrackCue} in Figures~\ref{sup_fig:flow_seflow}--\ref{sup_fig:flow_teflow} of Appendix~\ref{app:c}.

\begin{table}[t!]
\centering
\small
\setlength{\tabcolsep}{0.5em}
\renewcommand{\arraystretch}{1.08}
\caption{
Comparison of dynamic pseudo-label generation quality and downstream scene flow performance.
We compare the DUFOMap~\cite{duberg2024dufomap} and our \textsc{TrackCue} for auto-label generation of both SeFlow~\cite{zhang2024seflow} and SeFlow++~\cite{zhang2025himo} pipelines.
The ground-truth dynamic ratio is $7.12\%$.
}
\label{tab:dyn_cluster_comp}
\resizebox{\linewidth}{!}{
\begin{tabular}{llcccccc}
\toprule
Auto-Labeling Method & Method
& Pred. Dyn. Ratio (\%) & Accuracy (\%) $\uparrow$ 
& Precision (\%) $\uparrow$ 
& Recall (\%) $\uparrow$ 
& F1 Score $\uparrow$ 
& Dyn. Bucket $\downarrow$ \\
\midrule

\multirow{2}{*}{SeFlow~\citep{zhang2024seflow}}
& DUFOMap
& 20.93 & 82.26 & 24.62 & \textbf{72.35} & 36.74 & 0.372 \\
& \textsc{TrackCue}~(Ours)
& 14.80 & \textbf{87.37} & \textbf{31.38} & 65.22 & \textbf{42.37} & \textbf{0.352} \\
\midrule

\multirow{2}{*}{SeFlow++~\citep{zhang2025himo}}
& DUFOMap & 10.05 & 91.89 & 45.07 & \textbf{63.59} & 52.75
& 0.328 \\
& \textsc{TrackCue}~(Ours)
& 7.18 & \textbf{93.83} & \textbf{56.62} & 57.09 & \textbf{56.85}
& \textbf{0.298} \\
\bottomrule
\end{tabular}
}
\vspace{-1em}
\end{table}

\subsection{Comparison of Auto-labeling for Scene Flow Estimation}

In this section, we quantify the improvement in dynamic label fidelity using precision, recall, and F1 score, together with the Dynamic Bucket-Normalized EPE as the downstream metric.
Following the official evaluation protocol, ground-truth dynamic labels are obtained by thresholding the residual motion between the ground-truth and the rigid flow at $0.05\,\mathrm{m}$ against which the predicted dynamic points of each auto-labeling method are evaluated.
As shown in Table~\ref{tab:dyn_cluster_comp}, \textsc{TrackCue} significantly reduces false-positive dynamic predictions and improves precision, while its conservative filtering leads to lower recall.
%
%
Specifically, \textsc{TrackCue} improves the F1 score by 5.63 and 4.10 percentage points for SeFlow and SeFlow++, respectively. This indicates a better balance between filtering noisy labels and retaining true dynamic points. This trade-off is important, since false positives and false negatives affect the training objective differently. 
A false positive applies the Dynamic Chamfer and clustering terms to a stationary point and removes its static supervision. Such errors can be particularly harmful in driving scenes, where static points dominate and even a low false-positive rate can produce a substantial number of incorrect dynamic labels. In contrast, a false negative removes correct dynamic supervision from a moving point, but its effect can be partially compensated by other correctly labeled points from the same object.
The cleaner supervision also improves downstream scene flow estimation, reducing the Dynamic Bucket-Normalized EPE from 0.372 to 0.352 for SeFlow and from 0.328 to 0.298 for SeFlow++. 

\subsection{Ablation Studies}

\myparagraph{Analysis on Point Tracker.}
Table~\ref{tab:tracker_comparison} shows the effect of diverse point trackers on dynamic labeling quality. From this experiment, we draw the following insights. 
First, the points misclassified by the LiDAR-only method are consistently refined by visual motion cues, regardless of the choice of the point tracking model. Specifically, compared with DUFOMap, \textsc{TrackCue} improves precision by 5.61--10.68 percentage points and F1 score by 4.94--5.63 points.
Second, we find that TAPNext yields the highest precision but extremely lower recall under the same dynamic labeling criterion, indicating that its trajectories provide reliable yet less comprehensive motion evidence.
As a result, \textsc{TrackCue} with TAPNext drops more true dynamic points, which can incorrectly encourage zero-flow estimation and lead to lower scene flow performance. 
These observations also imply that a high-fidelity label is crucial for each self-supervised loss term to provide appropriate supervision.

\begin{table*}[t!]
\centering
\scriptsize
\setlength{\tabcolsep}{0.22em}
\renewcommand{\arraystretch}{0.88}

\begin{minipage}{0.62\textwidth}
\centering
\caption{Effect of point tracker choice.}
\label{tab:tracker_comparison}
\begin{tabular}{lcccc}
\toprule
Metric
& DUFOMap~\cite{duberg2024dufomap}
& CoTracker3~\cite{karaev2025cotracker3}
& TAPNext~\cite{zholus2025tapnext}
& AllTracker~\cite{harley2025alltracker} \\
\midrule
GT Dyn. Ratio
& \multicolumn{4}{c}{7.12\%} \\
\midrule
Pred. Dyn. Ratio (\%)
& 20.93 & 14.80 & 10.27 & 15.93 \\
Precision (\%) $\uparrow$ 
& 24.62 & 31.38 & \textbf{35.30} & 30.23 \\
Recall (\%) $\uparrow$ 
& \textbf{72.35} & 65.22 & 50.89 & 67.62 \\
F1 Score $\uparrow$
& 36.74 & \textbf{42.37} & 41.68 & 41.78 \\
Dyn. Bucket $\downarrow$
& 0.372 & \textbf{0.352} & 0.423 & 0.356 \\
\bottomrule
\end{tabular}
\end{minipage}
\hfill
\begin{minipage}{0.35\textwidth}
\centering
\caption{Effect of image resolution.}
\label{tab:resolution_ablation}
\setlength{\tabcolsep}{0.65em}
\begin{tabular}{lccc}
\toprule
Metric
& $786^2$
& $512^2$
& $256^2$ \\
\midrule
GT Dyn. Ratio
& \multicolumn{3}{c}{7.12\%} \\
\midrule
Pred. Dyn. Ratio (\%)
& 14.80 & 13.60 & 12.57 \\
Precision (\%) $\uparrow$ 
& 31.38 & 32.12 & \textbf{32.78} \\
Recall (\%) $\uparrow$ 
& \textbf{65.22} & 61.35 & 57.85 \\
F1 $\uparrow$
& \textbf{42.37} & 42.16 & 41.85 \\
Dyn. Bucket $\downarrow$
& \textbf{0.352} & 0.359 & 0.409 \\
\bottomrule
\end{tabular}
\end{minipage}
\end{table*}

\begin{table}[t!]
\centering
\footnotesize
\setlength{\tabcolsep}{0.65em}
\renewcommand{\arraystretch}{1.05}
\caption{Effect of input video clip length on dynamic clustering quality.}
\vspace{-0.5em}
\label{tab:clip_length_ablation}
\resizebox{0.9\linewidth}{!}{
\begin{tabular}{ccccc}
\toprule
\multicolumn{5}{c}{CoTracker3~\citep{karaev2025cotracker3}} \\
\midrule
Clip Length (\#Frames)
& Pred. Dynamic Ratio (\%) 
& Precision / Recall (\%) $\uparrow$ 
& F1 Score $\uparrow$
& Dyn. Bucket $\downarrow$ \\
\midrule
3
& 12.84 & \textbf{31.61} / 56.99 & 40.67 & 0.371 \\
6
& 14.80 & 31.38 / 65.22 & \textbf{42.37} & \textbf{0.352} \\
9
& 16.48 & 29.11 / \textbf{67.37} & 40.65 & 0.357 \\
12
& 16.54 & 28.35 / 65.90 & 39.64 & 0.373 \\
\bottomrule
\end{tabular}
}
\vspace{-2em}
\end{table}

\myparagraph{Image Resolution Sensitivity.}
Table~\ref{tab:resolution_ablation} reports dynamic clustering quality under different image resolutions. Experiments are done using CoTracker3~\cite{karaev2025cotracker3} as a point tracker.
High-resolution images preserve finer visual details and provide more reliable point trajectories, allowing more true dynamic points to be retained as dynamic.
In contrast, low-resolution images introduce larger projection errors and tracking ambiguities when 3D points are projected onto the 2D image plane.
As a result, some true dynamic points are missed, leading to lower recall.
Consistent with the previous results, imbalanced dynamic classification introduces misleading supervision for object motion status.

\myparagraph{Effect of Video Length.}
Table~\ref{tab:clip_length_ablation} reports the classification quality and downstream scene flow estimation performance.
We observe that short clips provide insufficient trajectory evidence, causing more true dynamic points to be missed and resulting in lower recall.
As the clip length increases, \textsc{TrackCue} predicts more points as dynamic, leading to a higher predicted dynamic ratio. Increasing the clip length initially improves recall by providing more temporal evidence, while overly long clips introduce tracking errors that offset this benefit.
However, longer clips can accumulate tracking drifts, occlusion noise, and viewpoint changes, introducing additional false-positive dynamic labels and reducing precision.
Therefore, since a clip length of $6$ provides a favorable precision--recall trade-off, we use it as the default setting in our main experiments. 

\begin{table}[t]
\centering
\caption{Effect of trajectory filtering. All settings use the ego-compensated residual, to which
the joint visibility mask and temporal voting are added progressively. Results are obtained on
the Argoverse 2 validation set~\cite{Argoverse2}, where the ground-truth dynamic ratio
is 7.12\%.}
\vspace{-0.5em}
\label{tab:component_ablation}
\setlength{\tabcolsep}{3.4pt}
\footnotesize
\begin{tabular}{ccc|cccc}
\toprule
CoTracker3~\cite{karaev2025cotracker3} Mask & Temporal Voting & Joint Visibility Mask &
Precision (\%) $\uparrow$ & Recall (\%) $\uparrow$ & F1 $\uparrow$ & Dyn. Bucket $\downarrow$ \\
\midrule
\checkmark &            &            & 24.98 & \textbf{71.02} & 36.96 & 0.368 \\
\checkmark &   \checkmark  &  & 31.28 & 65.21 & 42.28 & 0.359 \\
\checkmark & \checkmark & \checkmark & \textbf{31.38} & 65.22 & \textbf{42.37} & \textbf{0.352} \\
\bottomrule
\end{tabular}
\vspace{-1.5em}
\end{table}

\myparagraph{Analysis of Trajectory Filtering.} Point trackers already predict a per-frame visibility mask, but Table~\ref{tab:component_ablation} shows that relying on it alone is insufficient. This setting yields 24.98\% precision and 36.96 F1 score, leaving the DUFOMap labels it starts from essentially unchanged. Point tracking by itself therefore provides little benefit without the proposed filtering. 
To improve the reliability of the motion evidence, we introduce temporal voting and joint visibility filtering.
Temporal voting yields an F1 score of 42.28, raising precision to 31.28\% while retaining a recall of 65.21\%, since transient correspondence errors rarely persist across multiple frames. 
Further adding the joint visibility mask restricts residual computation to frames where both the tracked and rigid trajectories are valid, yielding 31.38\% precision and an F1 score of 42.37. It also reduces the Dynamic Bucket error from 0.359 to 0.352. These results show that the reliability of the visual motion cues comes from the proposed filtering rather than from the point tracker itself.

\begin{wrapfigure}{r}{0.45\textwidth}
\vspace{-1pt}
\centering
\includegraphics[width=0.45\textwidth]{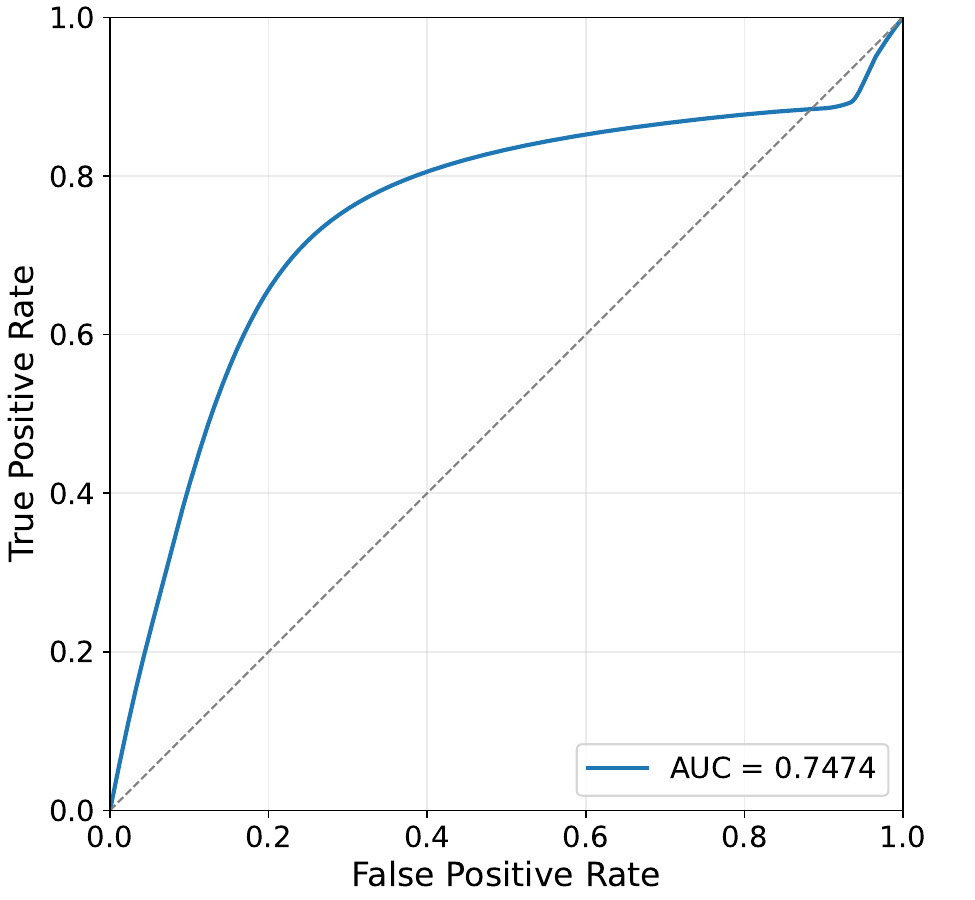}
\vspace{-1em}
\caption{ROC curve of the trajectory motion score. Sweeping $\tau_{\text{dyn}}$ over the
second-largest valid residual yields an AUROC of $0.7474$.}
\label{fig:roc_residual}
\vspace{-10pt}
\end{wrapfigure}
\myparagraph{Trajectory Residual Analysis.} \textsc{TrackCue} identifies dynamic queries by comparing tracked trajectories with those expected under ego-motion alone.
The residual in Eq.~\ref{eq:residual} captures motion that cannot be explained by ego-motion, and a query is classified as moving only when sufficiently large residuals persist over multiple jointly valid frames. 
To evaluate the discriminative ability of this residual independently of a particular threshold choice, we construct an ROC curve.
For each trajectory, we use the second-largest residual over jointly valid frames as the trajectory-level motion score, which is equivalent to the \(N_{\mathrm{move}}=2\) criterion in Eq.~\ref{eq:mask}. We vary \(\tau_{\mathrm{dyn}}\) over the range of trajectory-level scores to trace the ROC curve shown in Figure~\ref{fig:roc_residual}. 
The resulting AUROC of $0.7474$, well above the chance level of $0.5$, indicates that the trajectory residual provides meaningful discrimination between dynamic and static candidates independently of a particular choice of $\tau_{\mathrm{dyn}}$. This supports the interpretation that the precision gain is guided by discriminative motion evidence rather than an arbitrary reduction in the predicted dynamic ratio.
The curve rises steeply in the low false-positive regime, while the gain in the true-positive rate diminishes as the threshold is relaxed. This suggests that the trajectory residual is particularly effective for conservative dynamic detection, whereas further relaxing the threshold yields diminishing returns in recall.

\section{Conclusion}
We introduce \textsc{TrackCue}, a fully automated pipeline for refining static-dynamic pseudo-labels in self-supervised LiDAR scene flow estimation. We find that LiDAR-only dynamic-awareness mapping produces unreliable labels under sparsity and occlusion. To address this limitation, \textsc{TrackCue} complements sparse LiDAR geometry with rich visual motion cues while compensating for ego-induced image motion and filtering unreliable trajectories. Our experiments show that the resulting labels substantially improve dynamic-label precision and F1 score, leading to consistent gains in scene flow estimation across multiple self-supervised baselines. These results demonstrate the potential of image-based motion cues as complementary supervision for self-supervised LiDAR scene flow.

\myparagraph{Limitations \& Future Work.} While \textsc{TrackCue} shows promising results, several limitations raise important research questions. First, tracking over 3D-to-2D flattened points is inherently limited, since projection inevitably loses 3D geometric information. As a result, small or distant objects may be missed or overly refined. Thus, object masks from foundation segmentation models such as SAM~\cite{kirillov2023segment} could provide additional spatial constraints, preventing small or far-range dynamic objects from being overly suppressed. Second, the scope of our work is limited to generating high-fidelity pseudo-labels for clear self-supervised signals. Future work could further explore richer cross-modal information exchange, where RGB and LiDAR cues mutually refine both supervision signals and feature representations for 3D motion learning.

\section*{Acknowledgment}
This research was supported by Autonomous Driving Development Center, Hyundai Motor Company R\&D Division (85\%),
Culture, Sports and Tourism R\&D Program through the Korea Creative Content Agency grant funded by the Ministry of Culture, Sports and Tourism (International Collaborative Research and Global Talent Development for the Development of Copyright Management and Protection Technologies for Generative AI, RS-2024-00345025, 10\%),
the National Research Foundation of Korea(NRF) grant funded by the Korea government(MSIT) (RS-2025-00521602, 1\%),
Institute of Information \& communications Technology Planning \& Evaluation (IITP) grant funded by the Korea government(MSIT) (
No. RS-2019-II190079, Artificial Intelligence Graduate School Program(Korea University), 1\%; 
IITP-2025-RS-2024-00436857, 1\%; 
IITP-2026-RS-2026-25615490, AI Star Fellowship Support Program, 1\%),
and Advanced GPU Utilization Support Program funded by the Government of the Republic of Korea (Ministry of Science and ICT, 1\%).

\bibliographystyle{unsrt}
\bibliography{ref}

\newpage
\appendix

\section*{\Large Appendix}

\section{Experiment settings}
\label{app:a}

\subsection{Dataset Details}

In this section, we provide additional details on the Argoverse 2 leaderboard~\cite{li2025uniflowzeroshotlidarscene} test set, which comprises five different source datasets: Argoverse 2~\cite{Argoverse2}, Waymo~\cite{sun2020scalability}, nuScenes~\cite{caesar2020nuscenes}, TruckScenes~\cite{fent2024man}, and AEVAScenes~\cite{aevascenes}. To ensure a fair evaluation, the source identity of each scene is hidden by replacing dataset-specific information with anonymized random identifiers.

\myparagraph{Argoverse 2} is a widely adopted benchmark for LiDAR scene flow estimation. It provides annotated sensor data from 1,000 scenes, split into 700 training, 150 validation, and 150 test scenes.
Each frame is captured with seven ring cameras, two stereo cameras, and a point cloud acquired by two 32-beam LiDAR sensors.
The dataset primarily covers urban driving scenarios in U.S. cities.

\myparagraph{Waymo} is another large-scale benchmark in 3D perception for automative vehicles. It consists of 1,150 scenes with multimodal sensor data, including images from five high-resolution surround-view cameras and point clouds from five 64-beam LiDAR sensors. The dataset spans diverse driving conditions, ranging from urban and suburban environments to challenging nighttime scenarios.

\myparagraph{nuScenes} also contains 1,000 fully annotated scenes for various downstream 3D perception tasks and end-to-end autonomous driving. It captures urban driving scenarios using six surround-view cameras, a 32-beam LiDAR sensor, and five radar sensors.

\myparagraph{TruckScenes} contains geometric distribution shifts by collecting driving data from trucks, which differ from vehicles in both sensor placement and platform geometry.
It provides over 740 driving scenes, each lasting 20 seconds, across a broad range of environmental conditions.
Each scene primarily captures high-speed highway scenarios using four pinhole cameras, six 64-beam LiDAR sensors, and six radar sensors.

\myparagraph{AEVAScenes} provides high-fidelity LiDAR data with per-point measurements and long-range sensing capabilities for 3D scene understanding.
The dataset contains 10,000 frames captured by six wide- and narrow-FOV cameras, along with FMCW LiDAR sensors that cover front and rear views up to 200\,m.
It includes diverse driving conditions across cities and highways, covering both daytime and nighttime scenes in the Bay Area.

\subsection{Implementation Details}

In this work, we focus on self-supervised LiDAR scene flow estimation. We choose three strong baselines: SeFlow~(ECCV’24)~\cite{zhang2024seflow}, SeFlow++~(T-RO’25)~\cite{zhang2025himo}, and TeFlow~(CVPR’26)~\cite{zhang2026teflow}. Note that we use the official OpenSceneFlow repository\footnote{https://github.com/KTH-RPL/OpenSceneFlow} to reproduce all experimental results. Table~\ref{tab:training_config} summarizes the experimental configurations of the three baselines. Scene flow models with \textsc{TrackCue} labels are trained on four NVIDIA B200 GPUs, and all methods are evaluated on a single NVIDIA RTX A6000 GPU. The point tracker is used off-the-shelf and remains frozen throughout the pipeline.

\begin{table}[htb!]
\centering
\small
\setlength{\tabcolsep}{0.45em}
\renewcommand{\arraystretch}{1.08}
\caption{
Training configurations for self-supervised scene flow baselines. All configurations follow the official OpenSceneFlow implementation.
}
\vspace{-0.5em}
\label{tab:training_config}
\resizebox{\columnwidth}{!}{
\begin{tabular}{llccc}
\toprule
Category & Specification
& SeFlow~\cite{zhang2024seflow}
& SeFlow++~\cite{zhang2025himo}
& TeFlow~\cite{zhang2026teflow}\\
\midrule

\multirow{4}{*}{Model}
& Backbone
& DeFlow~\cite{zhang2024deflow}
& DeFlow~\cite{zhang2024deflow}
& DeltaFlow~\cite{zhang2026deltaflow} \\
& Input frames
& 2
& 3
& 5 \\
& Voxel size
& $[0.2,0.2,6]$
& $[0.2,0.2,6]$
& $[0.15,0.15,0.15]$ \\
& Point cloud range
& $[-51.2,-51.2,-3,51.2,51.2,3]$
& $[-51.2,-51.2,-3,51.2,51.2,3]$
& $[-38.4,-38.4,-3,38.4,38.4,3]$ \\
\midrule

\multirow{6}{*}{Training}
& Auto-labeler for SSL
& SeFlow auto-labeler
& SeFlow++ auto-labeler
& SeFlow auto-labeler \\
& Optimizer
& AdamW
& AdamW
& Adam \\
& Learning rate
& $2{\times}10^{-4}$
& $2{\times}10^{-4}$
& $2{\times}10^{-3}$ \\
& Scheduler
& --
& StepLR
& StepLR \\
& Epochs
& 9
& 9
& 15 \\
& Batch size
& 16
& 4
& 2 \\

\bottomrule
\end{tabular}
}
\vspace{-0.5em}
\end{table}

\section{Method Details}
\label{app:b}
\subsection{Dynamic-awareness Mapping}
To obtain self-supervised signals without manual annotations, existing methods commonly rely on occupancy-based dynamic-awareness mapping. 
DUFOMap~\cite{duberg2024dufomap} uses voxel grids and ray-casting to identify dynamic points based on temporal occupancy consistency. 
It marks ray-traversed voxels as void regions and classifies points later observed in those regions as dynamic. 
Formally, let $\mathcal{V}^{\mathrm{void}}_{<t}$ denote the set of voxels observed as void before time $t$. A point $\mathbf{p}_{i,t} \in \mathcal{P}_t$ is classified as dynamic if its voxel index $\mathrm{v}(\mathbf{p}_{i,t})$ conflicts with the previous void region:
\begin{equation}
    m_i = \mathbb{I}\left[\mathrm{v}(\mathbf{p}_{i,t}) \in \mathcal{V}^{\mathrm{void}}_{<t}\right],
\end{equation}
where $m_i=1$ indicates a dynamic point and $m_i=0$ indicates a static point. 
Then, the DUFOMap-based dynamic point set is defined as
$\mathcal{P}^{\mathrm{dufo}}_t = \{\mathbf{p}_{i,t} \mid m_i = 1\}$.
Although DUFOMap provides static-dynamic labels without manual annotations, it is sensitive to LiDAR sparsity and occlusions, which introduce noisy static-dynamic labels.

\subsection{Auto-labeling in self-supervised learning}
\myparagraph{SeFlow~\cite{zhang2024seflow} auto-labeler.} SeFlow first proposes an auto-labeling strategy in self-supervised methods that use the DUFOMap-based dynamic point set $\mathcal{P}^\mathrm{dufo}$ to construct self-supervision signals. While static points are constrained to have zero motion, dynamic points are clustered and optimized using geometric consistency objectives. To improve the accuracy of these dynamic labels, a cluster-level reassignment strategy is employed. Here, $\mathcal{P}^\mathrm{c}$ is obtained by applying HDBSCAN~\cite{campello2013density} only to dynamic points, with the minimum cluster size set to 20 and the cluster selection epsilon set to 0.7. Ultimately, $\mathcal{P}^\mathrm{c}$ is aligned with $\mathcal{P}^\mathrm{dufo}$ by masking out non-DUFO dynamic points. 

\myparagraph{SeFlow++~\cite{zhang2025himo} auto-labeler.} This strategy incorporates a nearest-neighbor dynamic point set $\mathcal{P}^\mathrm{nnd}$ into the SeFlow auto-labeler.
Specifically, a point $p \in \mathcal{P}_{t}$ is assigned to $\mathcal{P}^{\mathrm{nnd}}$ if its nearest-neighbor distance to $\mathcal{P}_{t+1}$ exceeds a user-defined threshold $\tau_d$:
\[
\mathcal{P}^{\mathrm{nnd}} =
\{p \mid D_{\min}(p, \mathcal{P}_{t+1}) > \tau_d,\; p \in \mathcal{P}_{t}\}, 
\]
where $D_{\min}$ computes the distance between a selected point $p$ and its nearest-neighbor counterpart. For Argoverse 2, they set the nearest-neighbor distance threshold to $\tau_d=0.14\,\mathrm{m}$. 
Given a set of clusters $\mathcal{C}$, each cluster $C \in \mathcal{C}$
is classified as dynamic based on the ratio of dynamic points:
\begin{equation}
f(C) =
\begin{cases}
\mathrm{Dynamic}, &
\text{if } \min(r_1,r_2) \geq \tau_1
\land \max(r_1,r_2) \geq \tau_2,\\
\mathrm{Static}, &
\text{otherwise},
\end{cases}
\end{equation}
where $r_1 =
\frac{|\mathcal{P}^{\mathrm{dufo}} \cap C|}{|C|}$, $r_2 =
\frac{|\mathcal{P}^{\mathrm{nnd}} \cap C|}{|C|}$. Following the given configuration, we set $\tau_1=0.05$ and $\tau_2=0.30$ for Argoverse 2.
Although this cluster-level reassignment improves robustness, the auto-labeling process still remains entirely dependent on LiDAR geometry. Consequently, the resulting labels are still vulnerable to errors caused by sparsity and occlusions.
Thus, starting from the DUFOMap-based dynamic point set $\mathcal{P}^\mathrm{dufo}$, our proposed \textsc{TrackCue} pipeline produces the refined dynamic point set $\mathcal{P}^{\textsc{TrackCue}}$. 
For all experiments, we replace $\mathcal{P}^{\mathrm{dufo}}$ with $\mathcal{P}^{\textsc{TrackCue}}$ when generating pseudo-labels for each auto-labeling strategy.

\subsection{\textsc{TrackCue} Details}

This section summarizes the proposed \textsc{TrackCue} pipeline and presents the hyperparameter settings used throughout our experiments. 
After generating $\mathcal{P}^{\textsc{TrackCue}}$, we train the three baseline models following the official training setup (see Table~\ref{tab:training_config}).
In our main experiments, we adopt CoTracker3 as a point tracker.
Seven surround-view RGB images are resized to $768 \times 768$, and a 6-frame video clip is used as input. Furthermore, we project up to 2,048 points from $\mathcal{P}^{\mathrm{dufo}}$ onto the image plane as initial dynamic point queries.
For the \textsc{TrackCue} pipeline, we use fixed thresholds throughout the entire process.
We set the moving criterion $\tau_{\mathrm{dyn}}$ to $5.0$ pixels, require dynamic evidence over at least $2$ frames for temporal voting, and use a minimum of $10$ points for moving retention.
Then, we lift the selected visual motion cues back to the LiDAR point cloud.
We use the top-$K_{\mathrm{nn}}$ nearest projected points with $K_{\mathrm{nn}}=4$ and a lifting radius $\tau_{\mathrm{lift}}=0.4$.
To enhance clarity and facilitate understanding, we provide a pseudo-code of
\textsc{TrackCue} pipeline in Algorithm~\ref{alg:trackcue}.

\begin{algorithm}[!htb]
    \caption{PyTorch-style Pseudocode for \textsc{TrackCue}}
    \label{alg:trackcue}
    \begin{lstlisting}[
        language=Python,
        basicstyle=\small\ttfamily,
        columns=fullflexible,
        keepspaces=true,
        showstringspaces=false,
        breaklines=true,
        escapeinside={(*@}{@*)}
    ]
# input:
# P, I: LiDAR point clouds and RGB images
# G: ego poses
# D: DUFOMap dynamic labels ((*@$\mathcal{P}^{\mathrm{dufo}}$@*))
# F: Point tracker (e.g., CoTracker3)

# Hyperparameters:
# (*@$\tau_{\mathrm{dyn}}$@*): motion threshold
# (*@$N_{\mathrm{move}}$@*): minimum moving frames
# (*@$N_{\mathrm{point}}$@*): minimum number of dynamic points
# (*@$\tau_{\mathrm{lift}}$@*): lifting radius
# (*@$K_{nn}$@*): top-(*@$K_{nn}$@*) nearest projected points for lifting

# output:
# Y: (*@\textsc{TrackCue}@*) dynamic labels ((*@$\mathcal{P}^{\text{\textsc{TrackCue}}}$@*))

def trackcue(P, I, G, D, F):
    # initialize refined dynamic labels
    Y = init_dynamic_masks(P)

    # make short clip for tracking
    for clip in split_into_clips(P, I, G, D):
        for v in camera_views:
            # query LiDAR points from DUFOMap dynamic labels
            Q = select_lidar_queries(
                clip.P[0], clip.D[0], camera_view=v
            )

            # Image-based Point Tracking (Sec. 3.3)
            q = lidar_to_image_projection(Q.xyz, camera_view=v)
            z_img, vis = F.track(clip.I[:, v], q)

            # Visually Consistent Motion Compensation (Sec. 3.4)
            z_rigid, rigid_vis = rigid_reproject(
                Q.xyz, clip.G, camera_view=v
            )

            e = torch.norm(z_img - z_rigid, dim=-1)
            votes = ((e > tau_dyn) & vis & rigid_vis).sum(dim=1)
            dynamic = votes >= N_move
            dynamic = dynamic & (dynamic.sum() >= N_point)

            # Visual Motion Cue Lifting (Sec. 3.5)
            Y = assign_tracks_to_lidar(
                Y, z_img[dynamic], clip.P, camera_view=v,
                radius=tau_lift, top_k=K_nn
            )

    return Y
    \end{lstlisting}
\end{algorithm}

\section{Additional Experimental Results}
\label{app:c}

\begin{table*}[t!]
\centering
\footnotesize
\setlength{\tabcolsep}{0.48em}
\renewcommand{\arraystretch}{1.12}
\caption{Experimental results on the Argoverse 2 \underline{leaderboard}
~\cite{li2025uniflowzeroshotlidarscene} -- \textbf{AV2}~\cite{Argoverse2}}
\label{tab:av2_test_av2}
\begin{tabular}{l c cccc ccccc}
\toprule
\multirow{3}{*}{Method}
& \multirow{3}{*}{\#Frames}
& \multicolumn{4}{c}{Three-way EPE (cm) $\downarrow$}
& \multicolumn{5}{c}{Dynamic Bucket-Normalized $\downarrow$} \\
\cmidrule(lr){3-6} \cmidrule(lr){7-11}
&
& Mean & FD & FS & BS
& Mean & CAR & OTHER & PED. & VRU \\
\midrule
SeFlow~\cite{zhang2024seflow}
& 2
& 4.76 & \textbf{10.93} & 2.23 & 1.12
& 0.3113 & 0.2365 & \textbf{0.2949} & 0.4406 & 0.2733 \\
\rowcolor{oursblue}
\ours
& 2
& \bestgain{4.66}{0.10}
& \badloss{10.97}{0.04}
& \bestgain{2.00}{0.23}
& \bestgain{1.02}{0.10}
& \bestgain{0.3070}{0.0043}
& \bestgain{0.2313}{0.0052}
& \badloss{0.3198}{0.0249}
& \bestgain{0.4073}{0.0333}
& \bestgain{0.2695}{0.0038} \\
\addlinespace[0.25em]
SeFlow++~\cite{zhang2025himo}
& 3
& \textbf{4.20} & 9.49 & \textbf{2.02} & \textbf{1.08}
& 0.2827 & 0.2023 & 0.3075 & \textbf{0.3867} & 0.2343 \\
\rowcolor{oursblue}
\ours
& 3
& \badloss{4.22}{0.02}
& \bestgain{9.14}{0.35}
& \badloss{2.35}{0.33}
& \badloss{1.16}{0.08}
& \bestgain{0.2781}{0.0046}
& \bestgain{0.2009}{0.0014}
& \bestgain{0.2865}{0.0210}
& \badloss{0.3959}{0.0092}
& \bestgain{0.2291}{0.0052} \\
\addlinespace[0.25em]
TeFlow~\cite{zhang2026teflow}
& 5
& 3.54 & 7.02 & 2.18 & 1.42
& 0.2019 & \textbf{0.1673} & 0.2208 & 0.2734 & \textbf{0.1461} \\
\rowcolor{oursblue}
\ours
& 5
& \bestgain{3.41}{0.13}
& \bestgain{6.88}{0.14}
& \bestgain{1.98}{0.20}
& \bestgain{1.36}{0.06}
& \bestgain{0.1966}{0.0053}
& \badloss{0.1712}{0.0039}
& \bestgain{0.2165}{0.0043}
& \bestgain{0.2477}{0.0257}
& \badloss{0.1511}{0.0050} \\
\bottomrule
\end{tabular}
\vspace{-0.5em}
\end{table*}

\subsection{Quantitative Results.}
We further provide additional quantitative results on the leaderboard, including detailed class-wise scores and Three-way End-Point Error across the Argoverse 2 data split.
As shown in Table~\ref{tab:av2_test_av2}, \textsc{TrackCue} consistently yields overall improvements in terms of object classes. This implies that the proposed method effectively identifies not only vehicles but also small and slow pedestrians, ultimately strengthening the network's generalization capability through highly refined supervision.

\subsection{Comparison with Monocular 4D Reconstruction.}
A natural alternative to \textsc{TrackCue} is to recover full 3D trajectories through monocular 4D reconstruction and derive dynamic regions from them. Table~\ref{tab:trace_anything} compares against Trace Anything~\cite{liu2026trace}, where dynamic regions are obtained from the temporal variance of reconstructed trajectories and transferred to LiDAR points. With ground-truth camera poses, which eliminate pose estimation errors, Trace Anything still over-predicts dynamic points at a ratio of 15.66 compared to the ground-truth ratio of 7.12, and achieves 38.37 F1, compared with 42.37 for \textsc{TrackCue}. When poses are instead estimated from the predicted point maps, F1 drops to 19.07 as pose errors propagate through the reconstructed trajectories. In contrast, \textsc{TrackCue} uses vehicle odometry, which is independent of image-based pose estimation and already required by the self-supervised baselines for ego-motion decomposition in Eq.~\ref{eq:flow_decomp}. 
Even with ground-truth poses, Trace Anything requires $7.2\times$ more computation and $3.3\times$ more memory than \textsc{TrackCue}.
These results indicate that reliable static-dynamic labels do not follow automatically from the monocular 4D reconstruction and that the process is inefficient.

\begin{table}[t!]
\centering
\footnotesize
\setlength{\tabcolsep}{0.6em}
\renewcommand{\arraystretch}{1.0}
\caption{Comparison with monocular 4D reconstruction on the Argoverse 2 validation
set~\cite{Argoverse2}. Trace Anything~\cite{liu2026trace} is evaluated with its
officially released weights without training or fine-tuning on Argoverse 2, following the same
protocol as \textsc{TrackCue}. Computational cost is measured per scene on a single RTX A6000 GPU.
The ground-truth dynamic ratio is 7.12\%.}
\label{tab:trace_anything}
\begin{tabular}{lccccccc}
\toprule
\multirow{2}{*}{Method} & Pred. Dyn. & Precision & Recall & F1 Score
& Dyn. Bucket & Time & Mem. \\
& Ratio (\%) & (\%) $\uparrow$ & (\%) $\uparrow$ & $\uparrow$
& $\downarrow$ & (s) $\downarrow$ & (GB) $\downarrow$ \\
\midrule
Trace Anything (GT pose) & 15.66 & 27.90 & 61.37 & 38.37 & 0.428 & 406.1 & 11.24 \\
Trace Anything (Est. pose) & 19.49 & 13.02 & 35.63 & 19.07 & 0.587 & 759.6 & 11.24 \\
\midrule
\textsc{TrackCue} (Ours) & \textbf{14.80} & \textbf{31.38} & \textbf{65.22} & \textbf{42.37}
& \textbf{0.352} & \textbf{56.6} & \textbf{3.45} \\
\bottomrule
\end{tabular}
\end{table}

\subsection{Predicted Flow Magnitude on Dynamic Points.}
\label{app:flow_magnitude}
We further investigate the predicted residual flow $\Delta\mathcal{F}$ in Eq.~\ref{eq:flow_decomp}, which remains after removing the ego-motion flow $\mathcal{F}^{\mathrm{ego}}$ from the scene flow and is the component directly
predicted by the model. We measure its magnitude on ground-truth dynamic points to examine whether improved pseudo-labels reduce static-like motion predictions. Ground-truth dynamic points are those whose residual motion exceeds
$0.05\,\mathrm{m}$ per frame, so a prediction below this value assigns static-like motion to a point that genuinely moves. 
As shown in Table~\ref{tab:flow_magnitude}, SeFlow~\citep{zhang2024seflow} predicts less than $0.05\,\mathrm{m}$ of residual motion for 39.5\% of ground-truth dynamic points, whereas \textsc{TrackCue} reduces this fraction to 34.7\%. The median predicted residual-flow magnitude also increases from $0.24\,\mathrm{m}$ to $0.35\,\mathrm{m}$. Together with the lower Dynamic Bucket error, these results show that the cleaner dynamic supervision provided by \textsc{TrackCue} reduces static-like predictions on truly moving points while improving dynamic scene-flow estimation.

\begin{table}[t!]
\setlength{\tabcolsep}{3pt}
\centering
\caption{Distribution of predicted residual-flow magnitudes on ground-truth dynamic points and Dynamic Bucket-Normalized EPE on the Argoverse 2 validation set. The four bins report the fraction of ground-truth dynamic points grouped by the predicted residual-flow magnitude. The median is computed from the predicted residual-flow magnitudes of all ground-truth dynamic points. Following the official scene flow evaluation protocol~\citep{khatri2024can}, ground-truth dynamic points are defined using a $0.05\,\mathrm{m}$ residual-motion threshold. Accordingly, predictions below $0.05\,\mathrm{m}$ are considered static-like.}
\label{tab:flow_magnitude}
\footnotesize
\begin{tabular}{lcccccc}
\toprule
& \multicolumn{4}{c}{Fraction of ground-truth dynamic points (\%)}
& \multirow{2}{*}{Median (m)}
& \multirow{2}{*}{Dyn. Bucket $\downarrow$} \\
\cmidrule(lr){2-5}
Method
& $<0.05\,\mathrm{m}$ $\downarrow$
& $0.05$--$0.2\,\mathrm{m}$
& $0.2$--$0.5\,\mathrm{m}$
& $\geq0.5\,\mathrm{m}$
& & \\
\midrule
SeFlow~\citep{zhang2024seflow}
& 39.5 & 8.7 & 14.6 & 37.2 & 0.24 & 0.372 \\
+ \textsc{TrackCue}
& \textbf{34.7} & 8.3 & 16.0 & 41.0 & 0.35 & \textbf{0.352} \\
\bottomrule
\end{tabular}
\end{table}

\subsection{Qualitative Results.}

\begin{figure}[b!]
\begin{center}
\includegraphics[width=\linewidth]{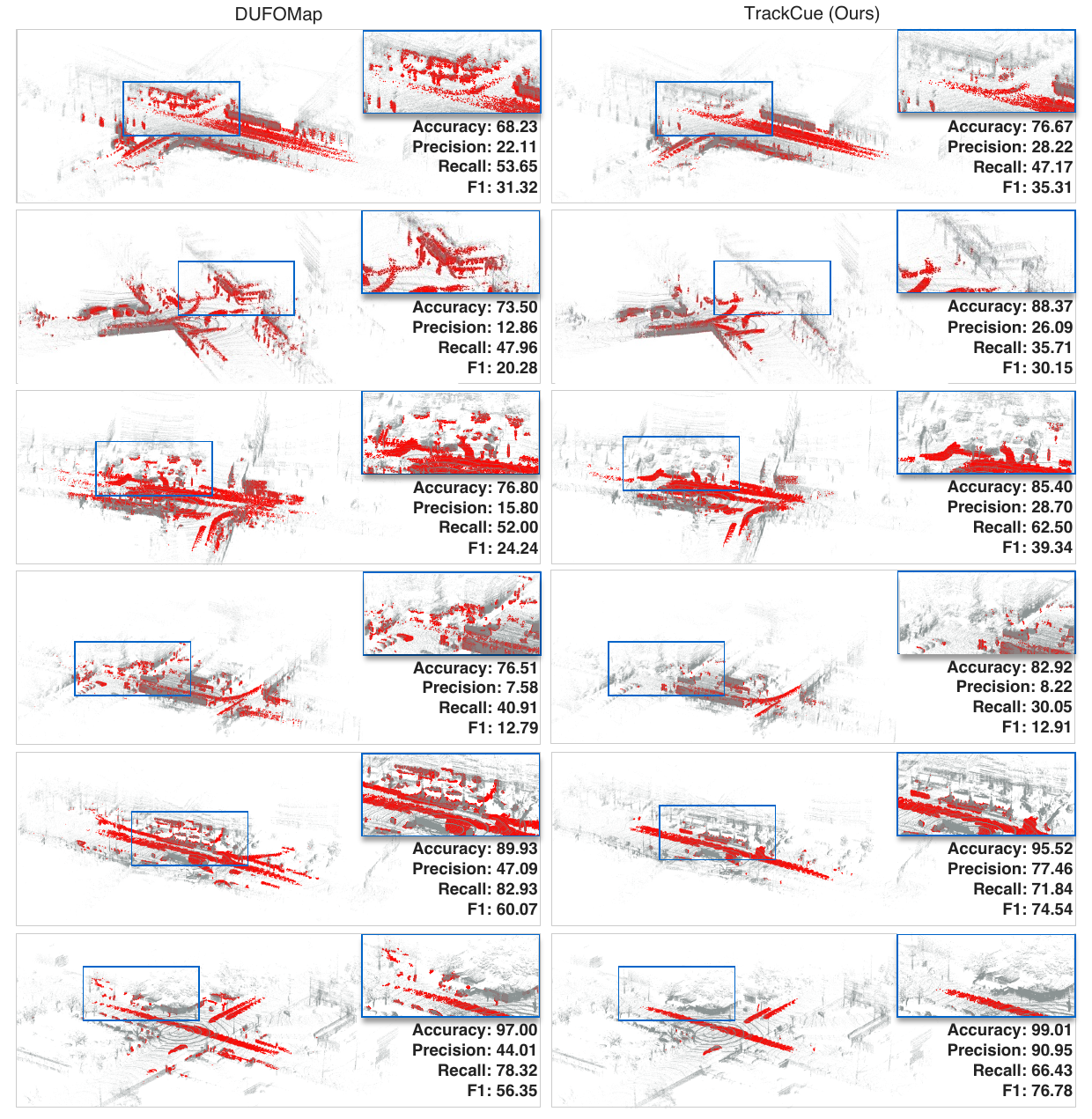}
\end{center}
\caption{Qualitative comparison of dynamic awareness map between DUFOMap~\cite{duberg2024dufomap} and \textsc{TrackCue}.}
\label{sup_fig:comp_dufo}
\end{figure}

\myparagraph{Comparison of Dynamic Clustering Quality.}
We visualize a dynamic awareness map by accumulating point-wise labels over a single scene. Red points denote non-stationary points. As clearly shown in Figure~\ref{sup_fig:comp_dufo}, \textsc{TrackCue} produces more consistent dynamic labels than DUFOMap~\cite{duberg2024dufomap}.
Importantly, we observe superiority with improvements of up to 46.94 percentage points in precision and 20.43 points in F1 score.

\begin{figure}[t!]
\centering
\includegraphics[
    width=\linewidth,
    height=0.9\textheight,
    keepaspectratio
]{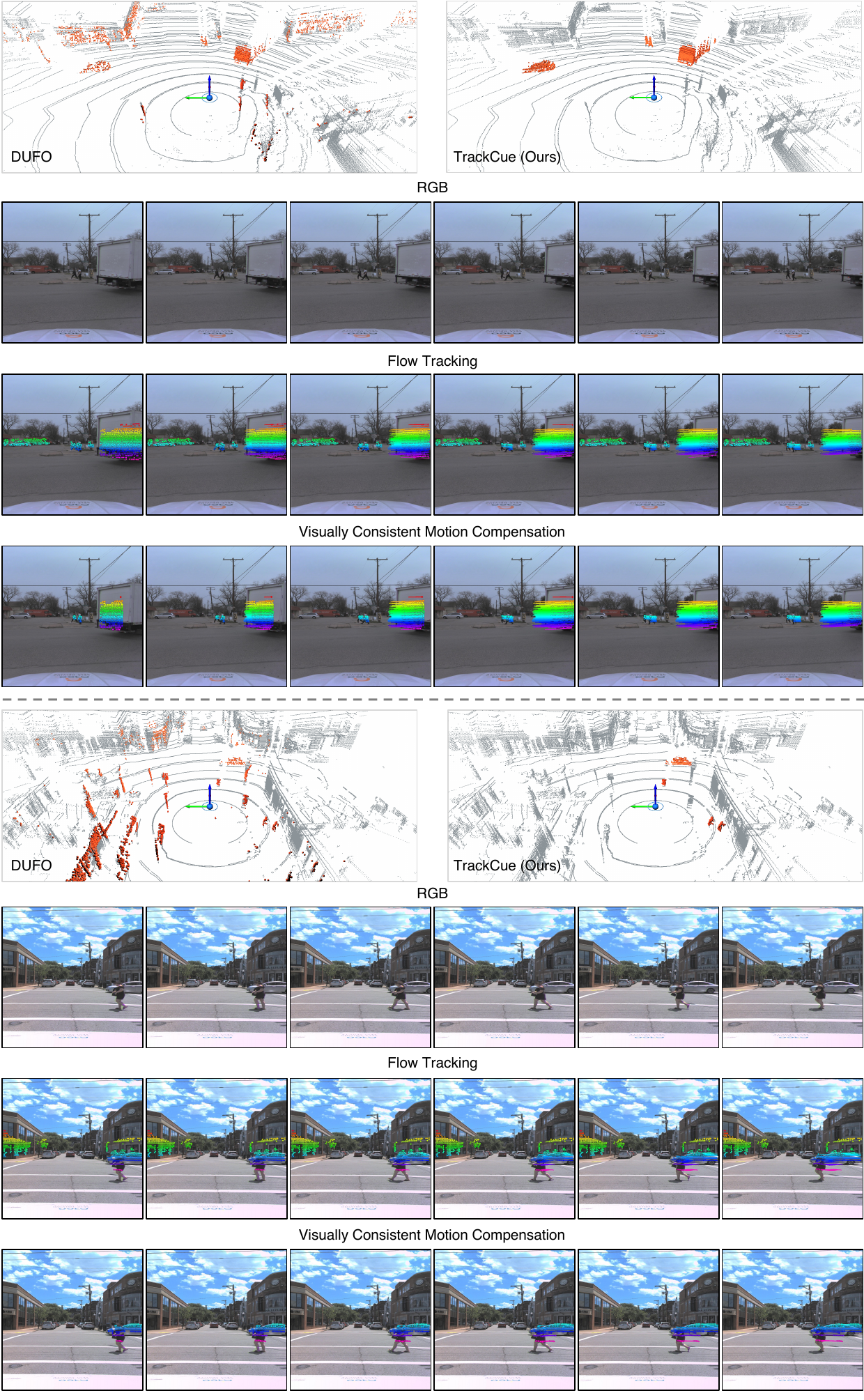}
\caption{Step-by-step visualization of \textsc{TrackCue}.}
\label{sup_fig:comp_flow1}
\end{figure}

\begin{figure}[t!]
\centering
\includegraphics[
    width=\linewidth,
    height=0.9\textheight,
    keepaspectratio
]{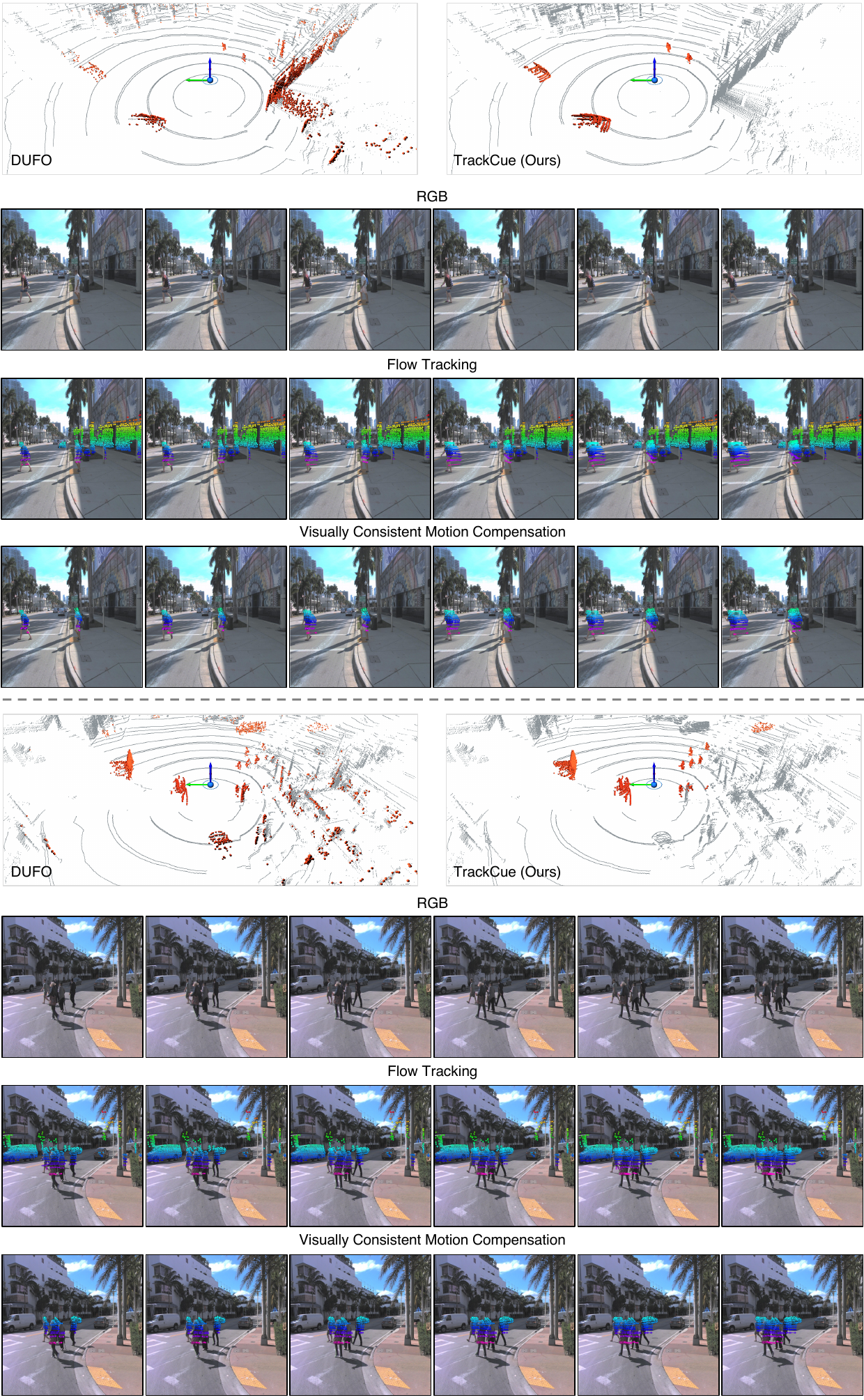}
\caption{Step-by-step visualization of \textsc{TrackCue}.}
\label{sup_fig:comp_flow2}
\end{figure}

\myparagraph{Visualization of \textsc{TrackCue} Results.}
Beyond numerical metrics, Figures~\ref{sup_fig:comp_flow1}--\ref{sup_fig:comp_flow2} further verify the effectiveness of \textsc{TrackCue} through step-by-step visualization results.
As shown in the top panel, the initial query points obtained from DUFOMap~\cite{duberg2024dufomap} are sparse and vulnerable to noise. Then, we scatter these points into image space, which is occupied by moving or non-moving objects. 
Using the point tracker~(e.g., CoTracker3~\cite{karaev2025cotracker3}), we acquire dense visual motion trajectories for candidate query points. 
After that, adjusting the rigid motion induced by the ego-vehicle successfully distinguishes static and dynamic objects. Specifically, it can capture pedestrian motions while rejecting temporarily stopped vehicles due to traffic conditions.
These ego-compensated 2D trajectories are projected back to 3D points by visual motion cue lifting. These results show that temporally consistent visual trajectories provide dense motion evidence, enabling \textsc{TrackCue} to complete initially partial dynamic labels.

\begin{figure}[t!]
\begin{center}
\includegraphics[width=\linewidth]{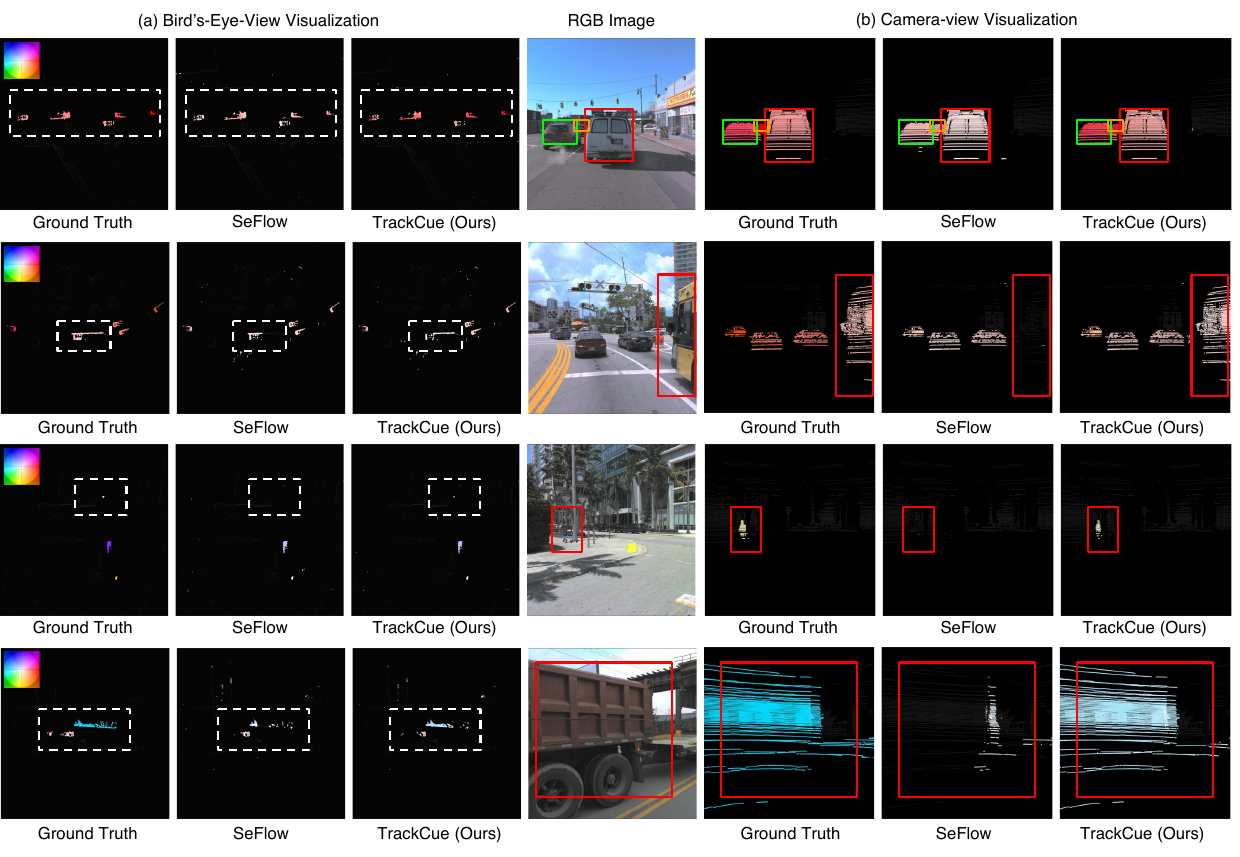}
\end{center}
\caption{Qualitative comparison with SeFlow~\cite{zhang2024seflow} for scene flow estimation.}
\label{sup_fig:flow_seflow}
\end{figure}

\myparagraph{Additional Qualitative Results for Scene Flow Estimation.}
For this study, we show two complementary views with corresponding images. The visualization results in Figures~\ref{sup_fig:flow_seflow}—\ref{sup_fig:flow_teflow} derive the following insights.
The baseline models suffer from the zero-flow estimation problem, producing weak motion responses for dynamic objects. For instance, the light hue in the color-coded flow indicates an underestimated motion magnitude. 
However, \textsc{TrackCue} provides reliable dynamic guidance while avoiding the bias toward zero-flow predictions, enabling scene flow models to learn more accurate motion through sophisticated self-supervised signals.
Second, \textsc{TrackCue} resolves the discontinuous labeling issue, where large dynamic objects, such as trucks, are partially classified as static.
Such cases are difficult to handle using sparse geometric cues alone, since the central regions of large objects can remain locally occupied across consecutive frames.
By leveraging dense visual trajectory cues, \textsc{TrackCue} propagates motion evidence across the object and produces more coherent dynamic labels.

\section{Broader Impact}
LiDAR scene flow estimation is an important component for autonomous driving, as it provides point-level motion cues that can support perception, prediction, and planning. By improving self-supervised scene flow learning without requiring dense ground-truth flow annotations, \textsc{TrackCue} can reduce the dependence on costly manual labeling and make motion estimation methods easier to scale to large driving datasets. The proposed use of image point tracking also highlights the potential of combining complementary sensor modalities to improve dynamic object understanding in sparse LiDAR scenes.
At the same time, the reliability of such systems remains critical. Since \textsc{TrackCue} uses image-based trajectories to refine dynamic labels, its performance may be affected by camera quality, illumination changes, adverse weather, calibration errors, and failures of the point tracker. Incorrect dynamic labels could propagate to scene flow learning and affect downstream autonomous driving modules. Therefore, additional validation under diverse driving conditions and careful integration with safety-critical perception pipelines are necessary before deployment in real-world autonomous systems.

\section{Code base and License}

Official URLs for the datasets and leaderboard utilized throughout our experiments~(see Table~\ref{tab:license_terms}).

\begin{table}[htbp]
\centering
\caption{License and URLs for Datasets}
\resizebox{0.95\columnwidth}{!}{%
\renewcommand{\arraystretch}{1.1}
\large
\begin{tabular}{c c c}
\toprule
\textbf{Name} & \textbf{License} & \textbf{URL} \\
\midrule
AV2 2026 Scene Flow Challenge  & Argoverse 2 Terms of Use & \url{https://www.argoverse.org/sceneflow.html} \\
OpenSceneFlow & BSD-3-Clause license & \url{https://github.com/KTH-RPL/OpenSceneFlow} \\
\bottomrule
\end{tabular}}
\label{tab:license_terms}
\end{table}

\begin{figure}[t!]
\begin{center}
\includegraphics[width=\linewidth]{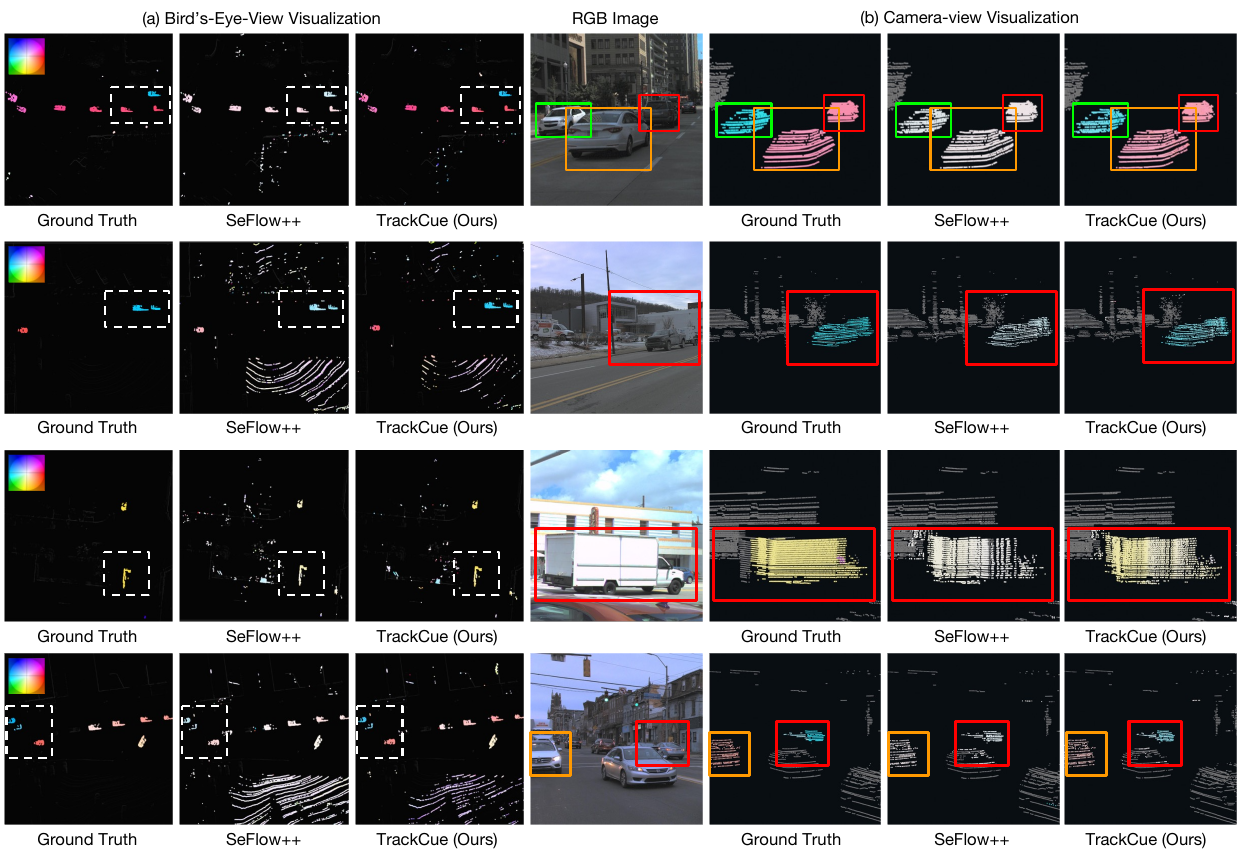}
\end{center}
\caption{Qualitative comparison with SeFlow++~\cite{zhang2025himo} for scene flow estimation.}
\label{sup_fig:flow_seflowpp}
\end{figure}

\begin{figure}[t!]
\begin{center}
\includegraphics[width=\linewidth]{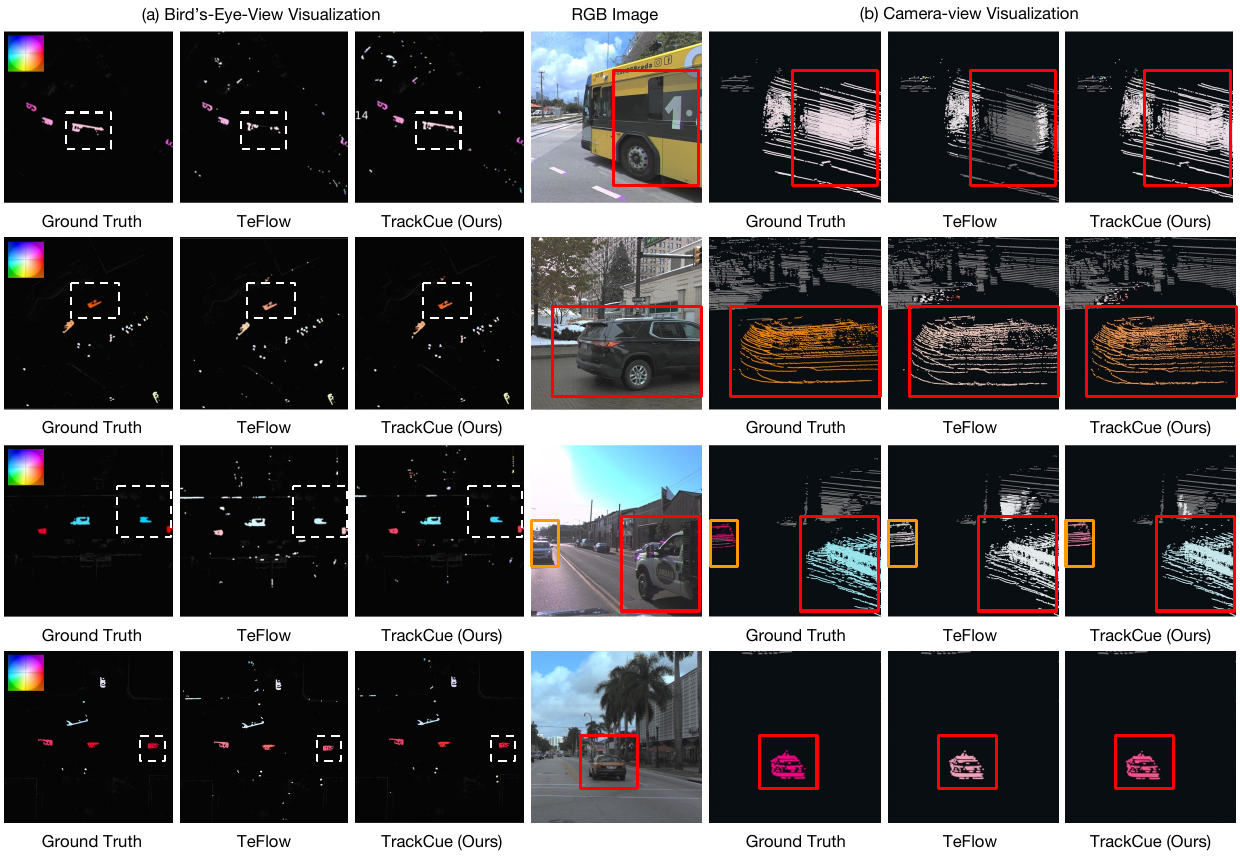}
\end{center}
\caption{Qualitative comparison with TeFlow~\cite{zhang2026teflow} for scene flow estimation.}
\label{sup_fig:flow_teflow}
\end{figure}


\end{document}